\documentclass[lettersize,journal]{IEEEtran}
\usepackage{amsmath,amsfonts}
\usepackage{array}
\usepackage[caption=false,font=normalsize,labelfont=sf,textfont=sf]{subfig}
\usepackage{textcomp}
\usepackage{stfloats}
\usepackage{url}
\usepackage{verbatim}
\usepackage{graphicx}
\usepackage{float}
\usepackage{booktabs}
\usepackage{multirow}
\usepackage{amsmath}
\usepackage{algorithm}
\usepackage{algpseudocode}
\usepackage{amssymb,amsmath,amsthm}

\usepackage{comment}
\usepackage{orcidlink}
\newtheorem{definition}{Definition}

\usepackage{cite}
\usepackage{amsmath,amssymb,amsfonts}
\usepackage{graphicx}
\usepackage{textcomp}
\usepackage{pifont}
\usepackage{comment}
\usepackage{multirow}
\usepackage{multirow}
\usepackage{booktabs}
\usepackage{adjustbox}
\usepackage{amsfonts}
\usepackage{amsmath}
\usepackage{algorithm}
\usepackage{algpseudocode}
\usepackage[switch]{lineno}
\usepackage{color}

\usepackage{newfloat}
\usepackage{listings}

\usepackage{amsthm}
\DeclareMathOperator{\softmax}{softmax}
\usepackage{enumitem} 
\usepackage{xcolor}

\usepackage{subcaption}  

\def\BibTeX{{\rm B\kern-.05em{\sc i\kern-.025em b}\kern-.08em
    T\kern-.1667em\lower.7ex\hbox{E}\kern-.125emX}}
\usepackage{balance}
\begin{document}
\title{Explainable Graph Neural Networks Under Fire}


\author{Zhong Li \orcidlink{0000-0003-1124-5778}, Simon Geisler \orcidlink{0000-0003-0867-1856}, Yuhang Wang \orcidlink{0009-0006-9085-6815}, Stephan Günnemann \orcidlink{0000-0001-7772-5059}, Matthijs van Leeuwen \orcidlink{0000-0002-0510-3549}\thanks{Zhong Li, Yuhang Wang, and Matthijs van Leeuwen are with Leiden University, the Netherlands. Simon Geisler, and Stephan Günnemann are with Technical University of Munich, Germany. Email: z.li@liacs.leidenuniv.nl (Zhong Li). (This work was primarily conducted while Zhong Li was a visiting PhD student with the DAML group at the
Technical University of Munich.)  }}

\markboth{Submitted to IEEE Transactions for possible publication}%
{How to Use the IEEEtran \LaTeX \ Templates}

\maketitle

\begin{abstract}
 Predictions made by graph neural networks (GNNs) usually lack interpretability due to their complex computational behavior and the abstract nature of graphs. 
 In an attempt to tackle this, many GNN explanation methods have emerged. Their goal is to explain a model's predictions and thereby obtain trust when GNN models are deployed in decision critical applications.
 Most GNN explanation methods work in a post-hoc manner and provide explanations in the form of a small subset of important edges and/or nodes. 
 In this paper we demonstrate that these explanations can unfortunately not be trusted, as common GNN explanation methods turn out to be highly susceptible to adversarial perturbations.
 That is, even small perturbations of the original graph structure that preserve the model's predictions may yield drastically different explanations. This calls into question the trustworthiness and practical utility of post-hoc explanation methods for GNNs. To be able to attack GNN explanation models, we devise a novel attack method dubbed \textit{GXAttack}, the first \textit{optimization-based} adversarial white-box attack method for post-hoc GNN explanations under such settings. Due to the devastating effectiveness of our attack, we call for an adversarial evaluation of future GNN explainers to demonstrate their robustness. For reproducibility, our code is available via \href{https://github.com/ZhongLIFR/GXAttack}{GitHub}.
\end{abstract}


\section{Introduction}
\label{sec:intro}
Graph Neural Networks (GNNs) \cite{zhou2020graph} have achieved promising results in various learning tasks, including representation learning \cite{chen2020graph, khoshraftar2024survey}, node classification \cite{rong2019dropedge, xiao2022graph}, link prediction \cite{kumar2020link, zhang2018link}, and anomaly detection \cite{li2024graphICSE,li2024cross}. However, intricate data representations and non-linear transformations render their interpretation and thus understanding their predictions non-trivial \cite{amara2022graphframex}.

This has led to the introduction of GNN explanation methods, which can improve a model's transparency and thus increase trust in a GNN model, especially when it is deployed in a decision-critical application (e.g., where fairness, privacy, or safety are important) \cite{ying2019gnnexplainer}. Moreover, they can help identify the scenarios where a GNN model may fail \cite{vu2020pgm}. Explainability of a GNN model can be obtained by designing a self-explainable GNN model, which usually employs a simple model architecture and few parameters, leading to explainability at the cost of suboptimal prediction accuracy; or by extracting post-hoc explanations that do not influence the internal workings of a GNN model and thus maintains its high prediction accuracy. In this paper we only consider post-hoc GNN explanation methods, as many such methods have merged over the past decade, including but not limited to \cite{ying2019gnnexplainer,pope2019explainability,yuan2020xgnn,luo2020parameterized,vu2020pgm,bajaj2021robust,yuan2021explainability,lucic2022cf,huang2022graphlime}. 


It has been shown that traditional deep neural networks (DNNs) are vulnerable to adversarial attacks \cite{szegedy2013intriguing,chakraborty2018adversarial}, where an imperceptible well-designed perturbation to input data can lead to substantially different prediction results. Recently, it was demonstrated that the explanations of predictions made by these DNNs are also fragile \cite{baniecki2024adversarial}. That is, an unnoticeable, well-designed perturbation to input data can result in completely different explanations while the model's predictions remain unchanged. However, the vulnerability of explanation methods has been primarily studied for DNN models designed for image and text data. In contrast, the vulnerability of GNN explanation methods to adversarial attacks has received very limited attention \cite{fang2024evaluating,wang2024v}, although a plethora of GNN explanation methods have recently been proposed \cite{yuan2022explainability,li2022survey2,kakkad2023survey}. 


Importantly, existing studies have only investigated the (in)stability of GNN explanations under relatively ``mild'' settings: they only consider \textit{random} perturbations and/or \textit{prediction-altering} perturbations.
Specifically, \cite{agarwal2022probing} theoretically analyzed the stability of three GNN explainers under random perturbations. Further, \cite{wang2024v} considered two scenarios: 1) graphs whose predictions are not changed after randomly flipping $20\%$ edges; and 2) severe perturbations using an off-the-shelf attack algorithm to perturb at most $10\%$ edges to change the GNN predictions (as well as the explanations). Besides, \cite{fan2023jointly} proposed an adversarial attack to simultaneously change the GNN predictions and their explanations. By altering predictions and explanations, however, the graph can become inherently different. \cite{fang2024evaluating} explored evaluation metrics for GNN explanations from the perspective of adversarial robustness and resistance to out-of-distribution input, where they defined ``adversarial robustness" as the difficulty of reversing a GNN prediction by perturbing the complementary of the explanatory subgraph.

\begin{figure}
    \centering
    \includegraphics[width=1\linewidth]{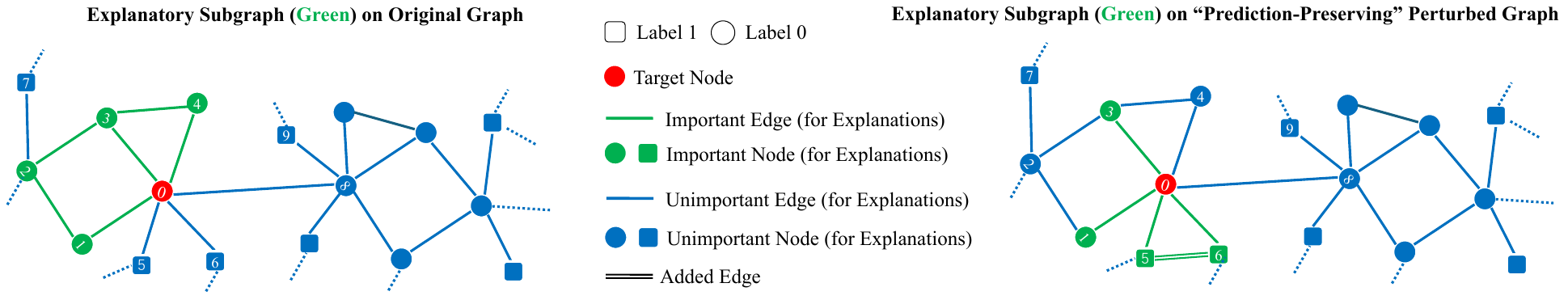}
    \caption{When using post-hoc GNN explainers, the explanatory subgraph on a graph with ``prediction-preserving" perturbations (right) can strongly differ from that on the original graph (left). }
    \label{fig:GXAttack_intro}
\end{figure}

Different from previous studies, we aim to perform perturbations that are both carefully crafted (i.e., \textit{optimized} rather than \textit{random})  and imperceptible (i.e., such that the GNN predictions remain unchanged but their explanations differ substantially).  As shown in Figure~\ref{fig:GXAttack_intro}, we empirically found that small \textit{prediction-preserving} perturbations can result in largely different explanations generated by post-hoc GNN explainers. To our knowledge, we are the first to formally formulate and systematically investigate this problem under a white-box setting. Specifically, we devise \textit{GXAttack}, the first \textit{optimization-based} adversarial white-box attack on post-hoc GNN explanations. We employ a widely used GNN explainer, PGExplainer \cite{luo2020parameterized}, as example target when designing our attack algorithm. Results on various datasets demonstrate the effectiveness of our approach. Moreover, our experiments show that other widely used GNN explanation methods, such as GradCAM \cite{pope2019explainability},  GNNExplainer \cite{ying2019gnnexplainer}, and SubgraphX \cite{yuan2021explainability}, are also fragile under the attacks optimized for PGExplainer. Due to the devastating effectiveness of our attack, we call for an adversarial evaluation of future GNN explainers to demonstrate their robustness. 



We first summarize related work in Section~\ref{sec:RelatedWork}. Following this,  we provide necessary background in  Section~\ref{sec:Preliminaries}. Then, we formalize the GNN explanation attack problem and introduce our novel attack algorithm GXAttack in Section~\ref{sec:Attack}. Next, we describe the experimental setups in Section~\ref{sec:Experiments}. Experiment results and corresponding analyses are given in Section~\ref{sec:ResultsAnlaysis}. Finally, we conclude the paper in Section~\ref{sec:conclusions}.


\section{Related Work}
\label{sec:RelatedWork}

\subsection{GNN Explanations}

\cite{yuan2022explainability, li2022survey2,kakkad2023survey} perform comprehensive surveys and propose excellent taxonomies of GNN explanation methods. Specifically, GNN explanations methods mainly include: 1) \textbf{Gradients-based methods}: Grad \cite{simonyan2013deep}, GradCAM \cite{pope2019explainability}, GuidedBP \cite{baldassarre2019explainability}, Integrated Gradients (IG) \cite{sundararajan2017axiomatic}; 2) \textbf{Decomposition-based methods}: GraphLRP \cite{schwarzenberg2019layerwise}, GNN-LRP \cite{schnake2021higher}; 3) \textbf{Surrogate-based methods}: GraphLIME \cite{huang2022graphlime}, PGM-Explainer \cite{vu2020pgm}; 4) \textbf{Generation-based methods}: XGNN \cite{yuan2020xgnn}; 5) \textbf{Perturbation-based methods}: GNNExplainer \cite{ying2019gnnexplainer}, PGExplainer \cite{luo2020parameterized}, GraphMask \cite{schlichtkrull2020interpreting}, SubgraphX \cite{yuan2021explainability}; and 6) \textbf{CF-based methods}: CF-GNNExplainer \cite{lucic2022cf}, RCExplainer \cite{bajaj2021robust}.

\subsection{Attacks and Defenses of Traditional XAI Methods}
It has been shown that traditional XAI methods (namely those designed to explain deep neural networks for image and text data) are susceptible to malicious perturbations \cite{heo2019fooling,zhang2020interpretable,ghorbani2019interpretation}. Please refer to \cite{baniecki2024adversarial} for a comprehensive survey.

\subsection{Robustness/Stability of GNN Explanations} \cite{yuan2022explainability,agarwal2022probing} point out that a reliable GNN explainer should exhibit stability, namely minor perturbations that do not impact the model predictions should not largely change the explanations. Particularly, \cite{agarwal2022probing} theoretically analyze the stability of three GNN explainers, including vanilla Grad, GraphMask and GraphLIME, for which they derive upper bounds on the instability of explanations. Importantluy, \cite{bajaj2021robust,fang2024evaluating,wang2024v,fan2023jointly} are closely related but are different from our work and they are detailed as follows.

\textbf{OAR.} \cite{fang2024evaluating} present a novel metric for evaluating GNN explanations, termed Out-Of-Distribution-resistant Adversarial Robustness (OAR). OAR aims to solve the limitations of current removal- and generative-based evaluations. More concretely, they evaluate post-hoc explanation subgraphs by computing their robustness under attack, which consists of three steps: 1) they formulate the adversarial robustness tailored for GNN explanations problem, which is the minimum adversarial perturbation on the structure of complementary subgraph;  2) they introduce a tractable and easy-to-implement objective of adversarial robustness for GNN explanations; after relaxation, the objective becomes a constrained graph generation problem (namely random perturbations) rather than an adversarial attack problem; 3) they present an Out-Of-Distribution (OOD) reweighting block that aims to confine the evaluation on original data distribution. This work is an initial attempt to explore evaluation metrics for GNN explanations from the perspectives of adversarial robustness and resistance to OOD. In contrast, we focus on optimization-based adversarial attack on
post-hoc GNN explanations rather than providing an evaluation metric.

\textbf{V-InFoR.} \cite{wang2024v} point out that ``existing GNN explainers are not robust to the structurally corrupted graphs, namely graphs with noisy or adversarial edges." This study investigates the negative effect of minor structural corruptions (which do not change the predictions) and severe structural corruptions (which change the predictions) on GNN explainers. Particularly, they provide quantitative evidence that existing GNN explainers are fragile to structurally corrupted graphs. They evaluate 6 GNN explainers under two settings: 1) minor corruptions where they select the perturbed graphs whose predictions are not changed after randomly flipping 20\% edges; 2) severe corruptions where they employ adversarial attack algorithm GRABNEL \cite{wan2021adversarial} to perturb at most 10\% edges to change the predictions.  In contrast, our method focus on optimization-based (rather than random) and prediction-preserving (rather than prediction-altering) adversarial attacks.

\textbf{RCExplainer.} \cite{bajaj2021robust} aim to find a small set of edges of the input graph such that the prediction result will substantially change if we remove those edges. Specifically, they propose RCExplainer to generate robust counter-factual explanations for GNNs in two steps: 1) they employ a set of decision regions to model the decision logic of a GNN, where each decision region determines the predictions on multiple graphs that are predicted to be the same class; 2) they leverage a DNN model to explore the decision logic, and thereby extract robust counter-factual explanations as a small set of edges of the input graph. This method is only applicable to GNNs that belong to Piecewise Linear Neural Networks. In this paper, we focus on attacking differentiable post-hoc GNN explanation methods that can be applied to any  GNNs.

\textbf{GEAttack.} \cite{fan2023jointly} empirically demonstrate that GNN explanation methods can be employed as tools to detect the adversarial perturbations on graphs. On this basis, they propose an attack framework dubbed \textit{GEAttack} that can jointly attack both GNN and its explanations. Specifically, they formulate \textit{GEAttack} as a bi-level optimization problem: 1) they mimic the GNN explanation method optimization process to obtain graph explanations in the inner loop; 2) they compute the gradient of the attack objective w.r.t. the explanations in the outer loop. Different from GEAttack, we consider prediction-preserving attacks, which are inherently more challenging.

\textbf{Concurrent Work.} While the concurrent work \cite{li2024graph} also addresses a similar topic, our study provides an alternative perspective by performing white-box attacks (namely we assume knowing the full knowledge about the GNN classifier and the GNN explainer), while they perform gray-box attacks by assuming knowing only the explanation loss and the generated explanatory edges of the GNN explainer.


\subsection{Adversarial Robustness of GNNs} Starting with the seminal works \cite{zugner2018adversarial,dai2018adversarial}, that both proposed adversarial attacks on GNNs, a rich literature formed in the realm of GNN-specific adversarial attacks, defenses, and certifications \cite{jin2021adversarial,gunnemann2022graph}. Most attacks primarily focus on altering predictions via perturbations of the discrete graph structure (edge insertions/deletions), either between existing nodes~\cite{zugner2018adversarial} or via the insertion of new (adversarial) nodes~\cite{zou_tdgiaeffective_2021}. Optimization-based attacks either operate globally (number of edge insertions/deletions)~\cite{xu2019topology,geisler2021robustness}, alter a local neighborhood~\cite{zugner2018adversarial}, or can handle a combination of both constraints~\cite{gosch_adversarial_2023}. While adversarial attacks on GNN explanation methods may benefit from optimization methods over the discrete graph structure, a careful discussion and derivation of appropriate attack objectives and metrics was missing.

\section{Preliminaries}
\label{sec:Preliminaries}
We utilize lowercase letters, bold lowercase letters, uppercase letters and calligraphic fonts to represent scalars ($x$), vectors ($\mathbf{x}$), matrices ($\mathbf{X}$), and sets ($\mathcal{X}$), respectively. 

\begin{definition}[Attributed Graph] We denote an attributed graph as $\mathcal{G} = \{\mathcal{V}, \mathcal{E}, \mathbf{X}\}$ where $\mathcal{V} = \{v_{1},...,v_{n}\}$ is the set of nodes. Let $\mathcal{E} = \{e_{ij}\}_{i,j \in\{1,...,n\}}$ be the set of edges, where $e_{ij}=1$ if there exists an edge between $v_{i}$ and $v_{j}$ and $e_{ij}=0$ otherwise. We use $\mathbf{A} \in \{0,1\}^{n\times n}$ to denote the corresponding adjacency matrix. $\mathbf{X} \in \mathbb{R}^{n\times d}$ represents the node attribute matrix, where the $i$-th row vector $\mathbf{x}_{i}$ denotes the node attribute of $v_{i}$. Thus, a graph can also be represented as $\mathcal{G} = (\mathbf{A}, \mathbf{X})$.
\end{definition}


\subsection{Node Classification with GNNs}

We consider the task of node classification. Given an input graph $\mathcal{G} = \{\mathcal{V}, \mathcal{E}, \mathbf{X}\}$ (or $\mathcal{G} = (\mathbf{A}, \mathbf{X})$), a prediction model $f_{\theta}(\cdot)$ 
assigns a label $c$ from a pre-defined set $\mathcal{C}$ to each node $v \in \mathcal{V}$. In other words, we have $f_{\theta}(\mathcal{G}(v)) = c$, where $\mathcal{G}(v)$ denotes the computation graph of node $v$. Typically, $f_{\theta}(\cdot)$ consists of two components: 1) a GNN model used to learn node representations, and 2) a multi-layer perceptron (MLP) classifier that assigns labels. More specifically, given a set of training nodes and their labels $\{(v_{1},y_{1}),...,(v_{n},y_{n})\}$ from graph $\mathcal{G}$, the objective function of a GNN classifier can be defined as
\begin{align}
    \underset{\theta}{\min}~\mathcal{L}_{\text{GNN}}(f_{\theta}) &:= \sum_{i=1}^{n}l(f_{\theta}(\mathcal{G}(v_{i}),y_{i}))\\
    &= \sum_{i=1}^{n}\sum_{c}^{C}\mathbb{I}(y_{i}=c)\ln(f_{\theta}(\mathcal{G}(v_{i}),y_{i})^{(c)}),
\end{align}
where $\theta$ is the set of learnable parameters for $f_{\theta}(\cdot)$, $l(\cdot)$ is the cross-entropy loss function, and $\mathbb{I}(\cdot)$ denotes the Kronecker function. Further, $f_{\theta}(\mathcal{G}(v_{i}),y_{i})^{(c)}$ indicates the $c$-th softmax output for the prediction of node $v_{i}$, and $C$ is the cardinality of label set $\mathcal{C}$.

Now we define the design detail of the GNN classifier $f_{\theta}(\cdot)$. For the first component, for simplicity, we employ a 2-layer GCN model with parameters $(\mathbf{W}_{1},\mathbf{W}_{2})$, i.e.,
\begin{equation}
    GCN(\mathcal{G}) = GCN(\mathbf{A},\mathbf{X}) := \Tilde{\mathbf{A}}\sigma(\Tilde{\mathbf{A}}\mathbf{X}\mathbf{W}_{1})\mathbf{W}_{2},
\end{equation}
where $\Tilde{\mathbf{A}} = \Tilde{\mathbf{D}}^{-1/2}(\mathbf{A}+\mathbf{I})\Tilde{\mathbf{D}}^{-1/2}$ (with $\mathbf{I}$ the identity matrix and $\Tilde{\mathbf{D}}$ the diagonal matrix of $\mathbf{A} + \mathbf{I}$). Further, $\sigma(\cdot)$ is an activation function such as ReLU. For the second component, we simply use a $\softmax(\cdot)$ function rather than a MLP. Therefore, the full GNN classifier $f_{\theta}$ is defined as
\begin{equation}
    f_{\theta}(\mathbf{A},\mathbf{X}) := \softmax(\Tilde{\mathbf{A}}\sigma(\Tilde{\mathbf{A}}\mathbf{X}\mathbf{W}_{1})\mathbf{W}_{2}).
\end{equation}

\subsection{GNN Explainer}

We consider post-hoc GNN explanation methods that provide instance-level explanations: given a graph $\mathcal{G}$, a target node $v$, and prediction $f_{\theta}(\mathcal{G}(v))$, the explainer $g_{\lambda}(v, f(\mathcal{G}(v)))$ aims to `explain' this particular prediction. Following common explanation methods, we assume that explanations are given in the form of an \textit{explanatory subgraph}, namely $\mathcal{G}_{s}(v) = g(v, f(\mathcal{G}(v)))$, consisting of a small set of important edges and their associated nodes, where the edge importance is measured by a so-called \textit{edge importance score}. 

In this paper, we take PGExplainer \cite{luo2020parameterized} as an example to attack for three reasons. First, \cite{faber2021comparing}  pointed out that GNN explanations providing edge importance are preferable to explanations providing node importance, due to the fact that edges have more fine-grained information than nodes. Second, \cite{agarwal2023evaluating} empirically showed that PGExplainer generates the least unstable explanations, exhibiting a $35.35\%$ reduction in explanation instability relative to the average instability of other GNN explainers. Third, the objective function of PGExplainer is differentiable. For completeness, we briefly revisit the goal of PGExplainer. For brevity, we denote $\mathcal{G}(v)$ as $\mathcal{G}$ (and $\mathcal{G}_{s}(v)$ as $\mathcal{G}_{s}$ analogously) throughout the remainder of the paper.

\textbf{PGExplainer.}
Given a graph $\mathcal{G}$ and a target node $v$, PGExplainer provides an \textit{explanatory subgraph} $\mathcal{G}_{s}(v) = (\mathbf{A}_{s},\mathbf{X}_{s})$ to explain why $f_{\theta}(\mathcal{G}(v))=c$, where $\mathbf{A}_{s}$ is the subgraph structure and $\mathbf{X}_{s}$ is the node features associated with nodes residing in the subgraph. Specifically, PGExplainer maximizes the mutual information between the GNN's prediction $\mathbf{y}$ and the explanatory subgraph $\mathcal{G}_{s}$, i.e.,
\begin{equation}
\label{equ:PGExplainerOriginal}
    \underset{\mathcal{G}_{s}}{\max}~\text{MI}(\mathbf{y},\mathcal{G}_{s}):=H(\mathbf{y})-H(\mathbf{y}|\mathcal{G}_{s}),
\end{equation}
where $H(\mathbf{y})$ is the entropy of $f_{\theta}(\mathcal{G})$, and $H(\mathbf{y}|\mathcal{G}_{s})$ is the conditional entropy of the GNN's prediction on the explanatory graph, namely $f_{\theta}(\mathcal{G}_{s})$. 
\section{GXAttack: \\ \underline{G}NN E\underline{x}planation Adversarial \underline{Attack}s}
\label{sec:Attack}

Given an input graph $\mathcal{G}$, a GNN classifier $f_{\theta}(\cdot)$, and a GNN explainer $g_{\lambda}(\cdot)$,  the original prediction for the target node $v$ is $f_{\theta}(\mathcal{G}(v))$ and its corresponding  original \textit{explanatory subgraph} is given by $\mathcal{G}_{s}(v) = g_{\lambda}(f_{\theta}(v,\mathcal{G}(v)))$. From an attacker's perspective, a desirable manipulated graph $\mathcal{G}_{adv} := \hat{\mathcal{G}} = \phi({\mathcal{G}})$ for attacking node $v$ should have the following properties,  where $\phi(\cdot)$ is an attack function that perturbs its input, i.e., it removes and/or adds a small set of edges:

\begin{enumerate}[align=left,leftmargin=*]
    \item The norm of the perturbation is so small that the change is considered imperceptible, i.e., $\mathcal{G} \approx \hat{\mathcal{G}}$. This can be formalized as $\Vert \mathbf{A} - \hat{\mathbf{A}} \Vert < B$, where $B$ is the perturbation budget;
    \item As GNNs can be highly sensitive to input perturbations, we require that the output of the GNN classifier remains approximately the same, i.e., perturbations must be \emph{prediction-preserving}. Formally, we have $f(\mathcal{G}(v)) = f(\hat{\mathcal{G}}(v))$. We hypothesize that perturbations that are both small and prediction-preserving are very likely to preserve the essence of the learned predictive models;
    \item The explanatory subgraph for node $v$ on original graph $\mathcal{G}_{s}(v) = g_{\lambda}(f_{\theta}(v,\mathcal{G}(v)))$ and that on perturbed graph  $\hat{\mathcal{G}}_{s}(v) = g_{\lambda}(f_{\theta}(\hat{\mathcal{G}}(v)))$  are substantially different: $\mathcal{G}_{s}(v)\neq\hat{\mathcal{G}}_{s}(v)$. In particular, we aim to make $\hat{\mathbf{A}}_{s}$ (i.e., the adjacency matrix of explanatory subgraph on perturbed graph) strongly different from $\mathbf{A}_{s}$ (i.e., the adjacency matrix of explanatory subgraph on original graph).
\end{enumerate}

\subsection{Attack Objectives}
\textbf{Generic Objective.} This paper focuses on adversarial structure attacks on post-hoc GNN explanation for node classification:
\begin{equation}
\label{FirstObjective}
 \underset{ \hat{\mathbf{A}} \text{~s.t.~} \Vert \mathbf{A} - \hat{\mathbf{A}} \Vert_{0} < B}{\max}~\mathcal{L}(g_{\lambda}(f_{\theta}(\hat{\mathbf{A}},\mathbf{X}))),
\end{equation}
with loss function $\mathcal{L}$, perturbation budget $B$, and  \textit{fixed} parameters $\theta$ and $\lambda$. The GNN predictor $f_{\theta}(\cdot)$ is applied to a graph $\mathcal{G}=\{\mathbf{A},\mathbf{X}\}$ to obtain node label predictions $f_{\theta}(\mathbf{A},\mathbf{X})$. Next, GNN explainer $g_{\lambda}(\cdot)$ is applied to GNN predictions $f_{\theta}(\mathbf{A},\mathbf{X})$ to generate post-hoc explanations $g_{\lambda}(f_{\theta}(\mathbf{A},\mathbf{X}))$. We focus on \textit{evasion} (test time) attacks and \textit{local} attacks on a single node with a budget $B$. Moreover, we study white box attacks as they represent the ``worst-case noise” of a model. In other words, we assume perfect knowledge about the graph, labels, prediction model, and post-hoc explanation model.

\textbf{Loss Function.} We should instantiate the loss function $\mathcal{L}$ such that it pertains to the desirable properties of manipulations discussed before. More specifically, it is defined as
\begin{align}
    \mathcal{L} :=~&\text{MI}(f_{\theta}(\mathbf{A},\mathbf{X}),f_{\theta}(\hat{\mathbf{A}},\mathbf{X}))\nonumber\\
    &- \beta* \text{MI}(g_{\lambda}(f_{\theta}(\mathbf{A},\mathbf{X})),g_{\lambda}(f_{\theta}(\hat{\mathbf{A}},\mathbf{X}))).
\end{align}
The first term ensures that the GNN predictions before and after perturbations are approximately the same. Further, the second term enforces that the manipulated explanation is different from the original explanatory subgraph. Their relative importance is weighted by hyperparameter $\beta \in \mathbb{R}_{+}$. 

\textbf{Final Objective.} We directly model the edge perturbation matrix $\mathbf{P}\in \{0,1\}^{n \times n}$, where we flip $\mathbf{A}_{ij}$ (namely $0 \to 1$ or $1 \to 0$) if $\mathbf{P}_{ij}=1$, and keep $\mathbf{A}_{ij}$ intact otherwise. On this basis, Eq~\ref{FirstObjective} can be rewritten as

\begin{align}
\label{SecondObjective}
 &\underset{ \mathbf{P} \in \{0,1\}^{n\times n} \text{~s.t.~} \sum\mathbf{P}_{ij} < B}{\max}~\text{MI}(f_{\theta}(\mathbf{A},\mathbf{X}),f_{\theta}(\mathbf{A}\oplus\mathbf{P},\mathbf{X}))\nonumber\\
 & - \beta* \text{MI}(g_{\lambda}(f_{\theta}(\mathbf{A},\mathbf{X})),g_{\lambda}(f_{\theta}(\mathbf{A}\oplus\mathbf{P},\mathbf{X}))),
\end{align}
where $\oplus$ means \textit{element-wise exclusive or} operation, and $B$ is the edge flipping budget. 

\subsection{Relaxations and Loss Definitions} Unfortunately, directly minimizing $\mathcal{L}$ w.r.t.\ $\mathbf{P}$ is intractable since there are $\mathcal{O}(2^{ n \times n})$ candidates for $\mathbf{P}$. We therefore approximate this discrete optimization problem, i.e.,
we make some relaxations to make $\mathcal{L}$ (approximately) differentiable w.r.t.\ $\mathbf{P}$ so that we can perform gradient-based attacks. 

\textbf{Relaxations.} We assume that edge flipping graph $\mathcal{M}$ corresponding to $\mathbf{P} \in \{0,1\}^{n \times n}$ (namely using $\mathbf{P}$ as the adjacency matrix) is a Gilbert random graph. In other words, edge occurrences are conditionally independent from each other. As a result, with slight abuse of notations, the probability of graph $\mathcal{M}$ can be factorized as $P(\mathcal{M})=\prod_{1\leq i, j \leq n}P(m_{ij})$, dubbed $\Tilde{\mathbf{P}}$. Next, we instantiate $P(m_{ij})$ using a Bernoulli distribution, namely $P(m_{ij}) \sim \text{Bernoulli}(\Tilde{\mathbf{P}}_{ij})$ with $P(m_{ij}=1)=\Tilde{\mathbf{P}}_{ij}$. As a result, Eq~\ref{SecondObjective} is equivalent to
\begin{equation}
\label{equ:PGExplainerRelax1}
    \underset{\mathbf{P}}{{\max}}~\mathcal{L}(\mathbf{P}) =\underset{\mathcal{M}}{\max}~\mathbb{E}_{\mathcal{M}}[\mathcal{L}(\mathcal{M})] \approx \underset{\Tilde{\mathbf{P}}}{\max}~\mathbb{E}_{\mathcal{M}\sim \Tilde{\mathbf{P}}}[\mathcal{L}(\mathcal{M})].
\end{equation}
In this way, we relax the $\mathbf{P} \in \{0,1\}^{n\times n}$ from binary variables to continuous variables $\Tilde{\mathbf{P}} \in [0,1]^{n\times n}$. Hence, we can utilise gradient-based methods to optimize the objective function. At the last step of the attack, we then discretize the result (see Section \ref{sec:optimization}).

\textbf{Loss Term 1.} We approximate the first term in $\mathcal{L}$ as follows:
\begin{align}
  &\underset{\Tilde{\mathbf{P}}}{\max}~ \text{MI}(f_{\theta}(\mathbf{A},\mathbf{X}),f_{\theta}(\mathbf{A}\oplus\Tilde{\mathbf{P}},\mathbf{X})) \\
  &:= \underset{\Tilde{\mathbf{P}}}{\max}~ H(f_{\theta}(\mathbf{A},\mathbf{X}))-H(f_{\theta}(\mathbf{A},\mathbf{X})|f_{\theta}(\mathbf{A}\oplus\Tilde{\mathbf{P}},\mathbf{X})) \label{MI:firstTerm1}\\
 &\propto  \underset{\Tilde{\mathbf{P}}}{\max}~ -H(f_{\theta}(\mathbf{A},\mathbf{X})|f_{\theta}(\mathbf{A}\oplus\Tilde{\mathbf{P}},\mathbf{X}))\label{MI:firstTerm2}\\
  &\approx  \underset{\Tilde{\mathbf{P}}}{\max}~ \sum_{c=1}^{C}P_{\theta}(y=c|\mathcal{G})\ln(P_{\theta}(y=c|\hat{\mathcal{G}})). \label{MI:firstTerm3}
\end{align}

Eq~\ref{MI:firstTerm1} holds by definition of mutual information;  Eq~\ref{MI:firstTerm2}  holds as $\theta, \mathbf{A}, \mathbf{X}$ are fixed;  Eq~\ref{MI:firstTerm3} holds when we  replace conditional entropy $H(f_{\theta}(\mathbf{A},\mathbf{X})|f_{\theta}(\mathbf{A}\oplus\Tilde{\mathbf{P}},\mathbf{X}))$ with cross entropy $H(f_{\theta}(\mathbf{A},\mathbf{X}), f_{\theta}(\mathbf{A}\oplus\Tilde{\mathbf{P}},\mathbf{X}))$ for practical optimization reason. Moreover, we define $\mathbf{A}_{ij} \oplus \Tilde{\mathbf{P}}_{ij}$ = $\mathbf{A}_{ij} + \Tilde{\mathbf{P}}_{ij}$ if $\mathbf{A}_{ij} =0$, and $\mathbf{A}_{ij}\oplus \Tilde{\mathbf{P}}_{ij}$ = $\mathbf{A}_{ij} - \Tilde{\mathbf{P}}_{ij}$ otherwise.

\textbf{Loss Term 2.} Next, we approximate the second term in $\mathcal{L}$ as follows:

\begin{align}
    &\underset{\Tilde{\mathbf{P}}}{\min}~ \text{MI}(g_{\lambda}(f_{\theta}(\mathbf{A},\mathbf{X})),g_{\lambda}(f_{\theta}(\mathbf{A}\oplus\Tilde{\mathbf{P}},\mathbf{X}))) \\
    & \propto \underset{\Tilde{\mathbf{P}}}{\max}~\text{Dist}( \mathbf{M}_{s},\hat{\mathbf{M}}_{s}) \label{MI:secondTerm3}\\
    &=:  \underset{\Tilde{\mathbf{P}}}{\max}~ \left(1-\frac{\text{vec}(\mathbf{M}_{pr}) \cdot \text{vec}(\hat{\mathbf{M}}_{pr})}{\Vert\mathbf{M}_{pr}\Vert_{F} \Vert\hat{\mathbf{M}}_{pr} \Vert_{F}}\right),
    \label{MI:secondTerm4}
\end{align}
where  $\mathbf{M} _ {s}$ and $\hat{\mathbf{M}} _ {s}$ are the explanatory subgraphs on original graph and perturbed graph, respectively; $\mathbf{M} _ {pr}$ and $\hat{\mathbf{M}} _ {pr}$ are the predicted explanations (edge importance matrices) on original graph and perturbed graph, respectively. In other words, the explanatory subgraph is represented as an edge importance matrix in this study. Eq~\ref{MI:secondTerm3} holds by definition; Eq~\ref{MI:secondTerm4} holds when we define the $Dist$ function as the \textit{cosine distance metric}, where $\text{vec}(\cdot)$ flattens a matrix to a vector, and $\Vert \cdot\Vert_{F}$ represents the Frobenius Norm.

\textbf{Final Relaxed Objective.} As a result, optimizing Eq~\ref{SecondObjective}  amounts to optimizing 
\begin{align}
\label{FinalObjective}
 \underset{ \Tilde{\mathbf{P}} \in [0,1]^{n\times n} \text{~s.t.~} \sum\Tilde{\mathbf{P}}_{ij} < B}{\max} &\sum_{c=1}^{C}P_{\theta}(y=c|\mathcal{G})\ln(P_{\theta}(y=c|\hat{\mathcal{G}})) \nonumber\\
 & + \beta*  \left(1-\frac{\text{vec}(\mathbf{M}_{pr}) \cdot \text{vec}(\hat{\mathbf{M}}_{pr})}{\Vert\mathbf{M}_{pr}\Vert_{F} \Vert\hat{\mathbf{M}}_{pr} \Vert_{F}}\right).
\end{align}

As a result, $\mathcal{L}(\Tilde{\mathbf{P}})$ (i.e., Eq~\ref{FinalObjective}) is differentiable w.r.t.~$\Tilde{\mathbf{P}}$.

\subsection{Optimization}\label{sec:optimization}


We employ Projected Gradient Ascent Algorithm (based on \cite{xu2019topology}) to optimize objective function~(\ref{FinalObjective}), and the details are given in Algorithm~\ref{algo:pga}. To go back from the continuous $\Tilde{\mathbf{P}}_{ij}$ to discrete ${\mathbf{P}}_{ij}$, we perform Bernoulli sampling with ${\mathbf{P}}_{ij} \sim \text{Bernoulli}(\Tilde{\mathbf{P}}_{ij})$ after optimization (Line~10). We note that the Projected Randomized Block Coordinate Descent (PR-BCD) \cite{geisler2021robustness} can perform this optimization more efficiently, but do not compare both methods due to the low scalability of the evaluated explainers.

\begin{algorithm*}[h]
      \small 
      \caption{Pseudo-code for GXAttack }
      \label{algo:pga}
      \begin{algorithmic}[1]
        \Require graph $\mathcal{G}=(\mathbf{A},\mathbf{X})$, predictor $f_{\theta}(\cdot)$, explainer $g_{\lambda}(\cdot)$, loss $\mathcal{L}$, budget $B$, \#epochs $T$, learning rate $\alpha_{t}$
        \Ensure perturbed adjacency matrix $\hat{\mathbf{A}}$
        \State \(\mathbf{y}\leftarrow f_{\theta}(\mathbf{A},\mathbf{X})\) \Comment{Get predictions on original graph}
        \State \(\mathcal{G}_{s} \leftarrow g_{\lambda}(f_{\theta}(\mathbf{A},\mathbf{X})\))\Comment{Get explanations for the predictions on original graph}
        \State $\Tilde{\mathbf{P}} \gets \mathbf{0} \in \mathbb{R}^{n \times n}$  \Comment{Initialize zeros for continuous perturbation variables }
        \For{$t \in \{1,2,...,T\}$}
        \State \(\hat{\mathbf{y}} \leftarrow f_{\theta}(\mathbf{A} \oplus\Tilde{\mathbf{P}}_{t-1},\mathbf{X})\) \Comment{Get predictions on perturbed graph}
        \State \(\hat{\mathcal{G}}_{s} \leftarrow g_{\lambda}(f_{\theta}(\mathbf{A} \oplus\Tilde{\mathbf{P}}_{t-1},\mathbf{X})\))\Comment{Get explanations for the predictions on perturbed graph}
        \State $\Tilde{\mathbf{P}}_{t} \gets \Tilde{\mathbf{P}}_{t-1} + \alpha_{t}\cdot\nabla_{\Tilde{\mathbf{P}}_{t-1}}\mathcal{L}(\hat{\mathbf{y}},\mathbf{y},\hat{\mathcal{G}}_{s}, \mathcal{G}_{s})$ \Comment{Update gradients}
         \State $\Tilde{\mathbf{P}}_{t} \gets \prod_{\mathbb{E}[\text{Bernoulli}(\Tilde{\mathbf{P}}_{t})\leq B]}(\Tilde{\mathbf{P}}_{t})$ 
        \Comment{Project gradients, see Appendix~\ref{app:PGD} for more detail}
        \EndFor
      \State $\mathbf{P}\sim\text{Bernoulli}(\Tilde{\mathbf{P}})$ s.t. $\sum\mathbf{P}_{ij} \leq B$ \Comment{Sample binary perturbation variables}
      \State \text{Return} $\mathbf{A}\oplus\mathbf{P}$ \Comment{Which is defined as $\mathbf{A}_{ij}+\mathbf{P}_{ij}(1-2\mathbf{A}_{ij})$}
      \end{algorithmic}
\end{algorithm*}

\subsection{Complexity Analysis}
We first discuss the dense case here and then make a discussion of GNNs/explainers implemented with sparse matrices. The time complexity is as follows:
\begin{enumerate}
    \item \textbf{Initial Operations}: $O(f_{\theta} + g_{\lambda})$, typically $O(n^2)$ for a dense GNN and explainer (sometimes even $O(n^3)$).
    \item \textbf{Each Epoch}:
    \begin{itemize}
        \item Prediction and explanation on perturbed graph: $O(2T \cdot (f_{\theta} + g_{\lambda}))$, where $T$ is the number of epochs.
        \item Gradient computation and update: Dependent on the efficiency of backpropagation through the GNN, but generally $O(T \cdot n^2)$ for gradient computation and a similar order for updates.
    \end{itemize}
    \item \textbf{Sampling Step}: $O(n^2)$.
\end{enumerate}
Meanwhile, the space complexity is as follows:
\begin{enumerate}
    \item \textbf{Space for Storing Matrices}: Storing $\mathbf{A}$, $\mathbf{X}$, $ \Tilde{\mathbf{P}}$ each requires $O(n^2)$ space.
    \item \textbf{Additional Space for Operations}: Space required for intermediate gradients and outputs during predictions, typically $O(n^2)$.
\end{enumerate}

Since the explainer uses dense matrices, we also choose an attack operating on all $n^2$ edges. Thus, the total complexity $ O((T + 1) \cdot (f_{\theta} + g_{\lambda}) + n^2)$ since we have $T$ attack steps plus the initial predictions as well as explanations and, last, we sample the final perturbations. Assuming that $ O(f_{\theta}) = O(g_{\lambda}) = O(n^2 d + n d^2)$ and $d \ll n$, this yields a total time complexity of $ O(n^2)$ with number of attack iterations $T \ll n$, and number of nodes $n$. This is the same asymptotic complexity as the dense GNN $f_{\theta}$ and explainer $g_{\lambda}$.

Assuming GNN and explainer are in $ O(f_{\theta}) = O(g_{\lambda}) = O(E d + n d^2)$ with number of edges $E$, a dense attack would dominate the overall complexity ($ O(n^2)$). An immediate remedy would yield PRBCD~\cite{geisler2021robustness} since its complexity is $O(B + E)$ with edge flipping budget $B$ and all additional components (e.g., loss function) could be implemented in $O(E)$ with sparse matrices. This yields an overall time and space complexity of $O(E)$, assuming $B \ll E$ and hidden dimensions $d \ll n$, which is the same asymptotic complexity as the sparse GNN $f_{\theta}$ and explainer $g_{\lambda}$.

\section{Experimental Setup}
\label{sec:Experiments}

We aim to answer the following research questions: (\textbf{RQ1})  How effective are the attacks generated by GXAttack on PGExplainer? (\textbf{RQ2}) How effective are the attacks generated by GXAttack when compared to trivial attackers that use random perturbations? and (\textbf{RQ3}) Can the attacks generated by GXAttack (optimized for PGExplainer) transfer to other GNN explanation methods?

\subsection{Datasets}

To answer these questions, we need to evaluate the GNN explanation quality before and after the attacks. However, as pointed out by \cite{agarwal2023evaluating}, evaluating the quality of post-hoc GNN explanations is challenging. This is because existing benchmark graph datasets lack ground-truth explanations or their explanations are not reliable. To mitigate this, they proposed a synthetic graph data generator \textit{ShapeGGen}. It can generate a variety of graph datasets with ground-truth explanations, where graph size, node degree distributions, the ground-truth explanation sizes, and more can be varied. As a result, \textit{ShapeGGen} allows us to mimic graph data in various real-world applications. More importantly, \textit{ShapeGGen} can ensure that the generated data and its ground-truth explanations do not suffer from GNN explanation evaluation pitfalls \cite{faber2021comparing}, namely trivial explanations, redundant explanations, and weak GNN predictors. This makes \textit{ShapeGGen} very suitable for studying the limitations of GNN explainers. We consider the synthetic datasets in Table~\ref{tab:syn_data_summary}.

\begin{table*}[]
	\caption{Summary of synthetic datasets with ground-truth explanations. \#Nodes and \#Edges represent the number of expected  nodes and edges, respectively. We do not consider larger datasets for two reasons: 1)  explanation accuracy will be very low ($< 0.3$) for most explainers, and it is not meaningful to attack low accuracy explainers; 2) computation will be slow for the attacker (and the explainer).}
	\centering
\resizebox{\linewidth}{!}{
	\begin{tabular}{p{2.3cm}p{1.2cm}p{1.2cm}p{1.2cm}p{1.2cm}p{1.2cm}p{1.2cm}p{1.2cm}}
		\toprule
		Dataset & Syn1 &Syn2 & Syn3& Syn4 & Syn5 & Syn6 & Syn7\\
		\midrule
        \textbf{Shape} & House & House & House & Circle & House & House & House\\
        \textbf{\#Subgraphs} & 10& 50 & 50 & 50 & 50 & 50 & 100\\
        \textbf{Subgraph Size} & 6 & 6 & 10 & 10 & 10 & 10 & 10\\ 
        \textbf{P(Connection)} & 0.3 & 0.06& 0.06 & 0.06 & 0.12 & 0.20 & 0.06\\ 
        \textbf{\#Classes} & 2 & 2 & 2 & 2 & 2 & 2 & 2\\
        \textbf{\#Nodes} & 59 & 299 & 521 & 511 & 489 & 492 & 1018\\
        \textbf{\#Edges} & 164 & 970 & 1432 & 1314 & 1664 & 1898 & 3374\\
        \textbf{Average Degree} & 2.8 & 3.2 & 2.7 & 2.6 & 3.4 & 3.9 & 3.3\\
        \textbf{Max Budget} & 10  & 15 & 15 & 15 & 15 & 15 & 15\\
        \bottomrule
	\end{tabular}
}
	\label{tab:syn_data_summary}
\end{table*}

\textbf{Why not real-world datasets?} We do not consider real-world datasets for two key reasons: 1) synthetic datasets with ground truth are \textit{de-facto} standards for evaluating post-hoc GNN explainers (for node classification), including GNNExplainer, PGExplainer, PGM-Explainer, SubgraphX, CF-GNNExplainer, RCExplainer, etc. 2) without ground-truth information we can only compute some metrics, such as cosine similarity of explanation before and after attack (which will be defined later) to quantify attack performance. However, as shown in Table~\ref{tab:algo_summary}, the cosine similarity and the real attack performance (in terms of $\Delta \text{GEA}$, which will be defined later) are not necessarily related. Therefore, considering such metrics may not be meaningful.


\subsection{Metrics and Baselines}
\textbf{Graph Explanation Accuracy (GEA).} We employ the \textit{graph explanation accuracy} proposed in \cite{agarwal2023evaluating} to quantify the quality of generated explanations  when the ground-truth explanations are available. For simplicity, we assume that each edge is binary coded as `0' (non-contributory) or `1' (contributory) to model predictions in both ground-truth (provided by \textit{ShapeGGen}) and predicted explanation masks (provided by the GNN explainer). Specifically, \textit{graph explanation accuracy} measures the correctness of generated explanations by computing the Jaccard Index between ground-truth explanation $\mathbf{M}_{gt}$ and predicted explanation $\mathbf{M}_{pr}$ as
\begin{align}
    &\text{JAC}(\mathbf{M}_{gt}, \mathbf{M}_{pr}) \nonumber\\
    &=\frac{\text{TP}(\mathbf{M}_{gt}, \mathbf{M}_{pr})}{\text{TP}(\mathbf{M}_{gt}, \mathbf{M}_{pr})+\text{FP}(\mathbf{M}_{gt}, \mathbf{M}_{pr})+\text{FN}(\mathbf{M}_{gt}, \mathbf{M}_{pr})},
\end{align}
where TP, FP, FN denote True Positives, False Positives, and False Negatives, respectively.  Higher values indicate larger consistency between the generated explanation and the ground-truth explanation. As a result, the difference in this metric before and after perturbation can be used to measure the performance of attacks: 
\begin{equation}
    \Delta \text{GEA} =: \text{JAC}(\mathbf{M}_{gt}, \mathbf{M}_{pr}) -  \text{JAC}(\mathbf{M}_{gt}, \hat{\mathbf{M}}_{pr}),
\end{equation} where higher values indicates more successful attacks.

\textbf{Cosine Similarity.} To quantify the performance of attacks when  ground-truth explanations are \textbf{not} available, we use cosine similarity to measure the stability of graph explanations:
\begin{equation}
    \operatorname{Sim}_{\cos}(\mathbf{M}_{pr}, \hat{\mathbf{M}}_{pr})=\frac{\text{vec}(\mathbf{M}_{pr}) \cdot \text{vec}(\hat{\mathbf{M}}_{pr})}{\Vert\mathbf{M}_{pr}\Vert_{F}\Vert\hat{\mathbf{M}}_{pr}\Vert_{F}},
\end{equation}
where $\text{vec}(\cdot)$ flattens its input as a vector, and $\Vert \cdot\Vert_{F}$ represents the Frobenius Norm. Higher values of $ \operatorname{Sim}_{\cos}$ (within $[-1,1]$) indicate higher graph explanation stability, and thus less successful attacks.

\textbf{Prediction Change.} To measure the impact of edge perturbations on GNN predictions, we consider two metrics as follows:  1) \textbf{$\Delta$Label}: the ratio of nodes of which the predicted label has changed after attacks; and 2) \textbf{$\Delta$Prob}: the average absolute change of predicted probability (of the original predicted class) of all nodes, which is formally defines as

\[
\Delta \text{Prob} = \frac{\sum_{i=1}^{N} \lvert P(y_{i}=y_{i}^{*}) - \hat{P}(y_{i}=y_{i}^{*}) \rvert}{N},
\]
where  \( P(y_{i}=y_{i}^{*}) \) is the prediction probability of node \( v_{i} \) being classified as \( y_{i}^{*} \) (the original predicted class with the highest probability) on the original graph. Moreover, \( \hat{P}(y_{i}=y_{i}^{*}) \) is the prediction probability of node \( v_{i} \) being classified as \( y_{i}^{*} \) (the original predicted class with the highest probability on the original graph) on the perturbed graph.

\textbf{Attacker Baselines.} Given the novelty of our problem setting, we only consider two trivial baselines for edge perturbations: 1) \textbf{random flipping}: we randomly flip $B_1$ edges; and 2) \textbf{random rewiring}: we randomly rewire $B_2$ edges within k-hop neighbours of the target node. 


\subsection{Attack Transferability Settings}

We  consider attack transferability on the following five types of GNN explainers: 1) \textbf{Gradient}-based: GradCAM \cite{pope2019explainability}, Integrated Gradients (IG) \cite{sundararajan2017axiomatic}; 2) \textbf{Perturbation}-based: GNNExplainer \cite{ying2019gnnexplainer}, SubgraphX \cite{yuan2021explainability}; 3) \textbf{Surrogate}-based: PGMExplainer \cite{vu2020pgm}; 4) \textbf{Decomposition-based methods}: GNN-LRP \cite{schnake2021higher} and 5) \textbf{Random} Explainer that randomly generates random important edges.

\subsection{Training Protocols and Compute Resources}
\label{app:trainingProtocols}
\textbf{GNN Predictor Training Protocols.} To avoid pitfalls caused by weak GNN predictor \cite{faber2021comparing}, we should train the GNN predictor close to its maximum possible performance. When training the GNN predictor, we employ an Adam optimizer where the learning rate is 1e-2, weight decay is 1e-5, the number of epochs is 300, and the hidden dimension is 32. 

\textbf{GNN Explainer Training Protocols.}
We followed the authors' recommendations for setting the hyperparameters of GNN explanation methods. We select  top-$k~(k=25\%)$ important edges to generate explanations for all GNN explainers. Particularly, when training the PGExplainer, the number of training epochs is 10. 

\textbf{GNN Explanation Attacker Training Protocols.}
We set the trade-off hyperparameter in the final loss as $0.1$ such that the magnitudes of the two loss terms are approximately the same. Besides, the learning rate in gradient ascent is set to $0.1$. Importantly, we set the maximal training epochs as 100 due to computation time reason, while further increasing this number can largely improve the attack performance (in terms of $\Delta$GEA). Moreover,  we set the perturbation budget to 15 at maximum for synthetic data (note that we attempt to employ a hard minimum budget of 5 with 2000 maximum trials of Bernoulli sampling). We perform sensitivity analysis regarding these hyperparameters, and the results and analysis will be discussed in Section~\ref{subsec:SenAna}.

\textbf{Explanation Accuracy Computation.} The $get\_acc(\cdot)$ function of GraphXAI considers all non-zero edges as predicted positives (namely with binary code `1') by default when computing JAC. Following \cite{agarwal2023evaluating}, we set the top 25\% (of edges with the highest edge importance scores) as 1 and the rest as zeros in our evaluations when computing both GEA and cosine similarity.

\textbf{Software and Hardware.}
The evaluated GNN explainers are implemented mainly based on GraphXAI \footnote{https://github.com/mims-harvard/GraphXAI} \cite{agarwal2023evaluating}. Algorithms are implemented in Python 3.8 (using PyTorch \cite{paszke2019pytorch}
and PyTorch Geometric \cite{fey2019fast} libraries when applicable) and ran on internal clusters with AMD 128 EPYC7702 CPUs (with 512GB RAM). All experiments can be finished within 7 days if using 10 such machine instances. All code and datasets are available on  GitHub\footnote{\url{https://github.com/ZhongLIFR/GXAttack}}.

\section{Experimental Results and Analysis}
\label{sec:ResultsAnlaysis}

Before answering the research questions, we state three presumptions and check them empirically in Section~\ref{subsec:presumptions}. Next, we answer RQ1 and RQ2 in Section~\ref{subsec:RQ12}, and RQ3 in Section~\ref{subsec:RQ3}. Following this, we perform sensitivity analysis in Section~\ref{subsec:SenAna} and property analysis in Section~\ref{subsec:ProAna}.

\subsection{Experiments to Check Presumptions}
\label{subsec:presumptions}

\textbf{Presumption 1.} \textit{The prediction correctness of a GNN predictor and explanation accuracy of a GNN explainer are \textbf{not} related.} For each dataset, we first get the set of correctly predicted nodes and the set of wrongly predicted nodes; then we compare the distribution of their explanation accuracy. As shown in Figure~\ref{fig:ExpAccVSPreAcc_All} in Appendix, there is no notable difference. In other words, we do \textbf{not} observe that wrongly labelled nodes tend to have lower explanation accuracy despite \cite{faber2021comparing} claim that it is unfair to use wrongly labelled nodes to evaluate GNN explanations (as they may not use the ground-truth explanations to make predictions and thus tend to have lower explanation accuracy).

\textbf{Presumption 2.} \textit{The prediction confidence of a GNN predictor and explanation accuracy of a GNN explainer are related,} where prediction confidence is defined as the probability assigned to the predicted class. For each dataset, we first divide the nodes into five groups based on their prediction confidence; then we compare the distribution of their explanation accuracy. Table~\ref{tab:PreConfidenceAll} in Appendix show that group `G9' (i.e., the set of nodes with prediction confidence in $[0.9, 1.0]$) indeed tends to have lower explanation accuracy on the original graph than other groups.

\textbf{Presumption 3.} \textit{The prediction confidence of a GNN predictor and attack performance of a GNN explanation attacker are related.} To check this presumption, we first divide the nodes into five groups based on their prediction confidence; then we compare the distribution of their attack performance. Table~\ref{tab:PreConfidenceAll} in Appendix show that group `G9'  also tends to have less successful attacks (i.e., smaller $\Delta \text{GEA}$) when compared to other groups. Possible reasons are that: 1) the nodes in this group have lower explanation accuracy on original graph; and 2) the prediction confidence is close to 1, leaving no much room for perturbations. These two facts together make the attack more challenging.

\subsection{Experiments to Answer RQ1 and RQ2: Attack Effectiveness}
\label{subsec:RQ12}
\begin{table*}[]
	\caption{Results on all synthetic datasets (average $\pm$ standard deviations across 5 trials). \textbf{O. GEA} means explanation accuracy on original graph and \textbf{P. GEA} indicates explanation accuracy on perturbed graph. Moreover, \textbf{$\Delta$GEA} is the explanation accuracy change after perturbation, \textbf{Sim}$_{\cos}$ is the cosine similarity between explanations (before and after perturbations), and \textbf{\#Pert.} represents the number of flipped edges. \textbf{$\Delta$Label} is the average change of predicted labels, while \textbf{$\Delta$Prob} denotes the average absolute change of predicted probability (of the original predicted class). ``GXAttack", ``Rnd. Flipping", and ``Rnd. Rewiring" correspond to our attack method, random flipping, and random rewiring, respectively. $`\uparrow'$ indicates larger values are preferred while $`\downarrow'$ means the opposite.}
	\centering
\resizebox{\linewidth}{!}{
\begin{tabular}{clrrrrrrr}%
    \toprule
     & \textbf{Dataset} & Syn1 &Syn2 & Syn3 & Syn4 & Syn5 & Syn6 & Syn7\\
    \midrule
    & \textbf{O. GEA} & $0.678_{\pm \text{0.046}}$  & $0.641_{\pm \text{0.022}}$ & $0.494_{\pm \text{0.012}}$ & $0.469_{\pm \text{0.006}}$ & $0.439_{\pm \text{0.014}}$ &  $0.306_{\pm \text{0.025}}$ &  $0.429_{\pm \text{0.013}}$\\
    \hline
    \multirow[c]{6}{*}{\rotatebox{90}{GXAttack}}& \textbf{P. GEA}$(\downarrow)$ & $0.497_{\pm \text{0.041}}$  & $0.375_{\pm \text{0.011}}$ & $0.308_{\pm \text{0.010}}$ & $0.312_{\pm \text{0.003}}$ & $0.290_{\pm \text{0.004}}$ &  $0.217_{\pm \text{0.013}}$ & $0.287_{\pm \text{0.014}}$\\
    &  \textbf{$\Delta$GEA} $(\uparrow)$ & $0.181_{\pm \text{0.020}}$ & $0.267_{\pm \text{0.014}}$ & $0.186_{\pm \text{0.013}}$ & $0.157_{\pm \text{0.010}}$ & $0.149_{\pm \text{0.015}}$ &  $0.088_{\pm \text{0.014}}$ & $0.142_{\pm \text{0.009}}$\\
    & \textbf{Sim}$\boldsymbol{_{\cos}}(\downarrow)$ &$0.921_{\pm \text{0.007}}$ & $0.975_{\pm \text{0.001}}$ & $0.980_{\pm \text{0.004}}$ & $0.976_{\pm \text{0.003}}$ & $0.978_{\pm \text{0.002}}$ &  $0.988_{\pm \text{0.001}}$ & $0.971_{\pm \text{0.040}}$\\
    & \textbf{\#Pert.} ($\downarrow$) &$3.300_{\pm \text{0.255}}$ & $5.620_{\pm \text{0.045}}$ & $6.680_{\pm \text{0.148}}$ & $5.900_{\pm \text{0.122}}$ & $6.560_{\pm \text{0.055}}$ &  $5.940_{\pm \text{0.167}}$ & $7.020_{\pm \text{0.130}}$\\
    & \textbf{$\Delta$Label}($\downarrow$) & $0.000_{\pm \text{0.000}}$ & $0.002_{\pm \text{0.003}}$ & $0.001_{\pm \text{0.001}}$  & $0.001_{\pm \text{0.001}}$ & $0.001_{\pm \text{0.001}}$ &  $0.000_{\pm \text{0.001}}$ & $0.001_{\pm \text{0.001}}$\\
    & \textbf{$\Delta$Prob}($\downarrow$) & $0.106_{\pm \text{0.008}}$& $0.140_{\pm \text{0.004}}$ & $0.163_{\pm \text{0.003}}$& $0.150_{\pm \text{0.002}}$ & $0.143_{\pm \text{0.004}}$ & $0.089_{\pm \text{0.003}}$ & $0.147_{\pm \text{0.001}}$\\
    \hline
    \multirow[c]{6}{*}{\rotatebox{90}{Rnd. Flipping}} & \textbf{P. GEA}$(\downarrow)$ &  $0.640_{\pm \text{0.056}}$ & $0.638_{\pm \text{0.023}}$ & $0.493_{\pm \text{0.012}}$ &  $0.467_{\pm \text{0.009}}$ & $0.438_{\pm \text{0.013}}$ &  $0.306_{\pm \text{0.024}}$ & $0.429_{\pm \text{0.013}}$\\
    & \textbf{$\Delta$GEA}$(\uparrow)$ & $0.038_{\pm \text{0.029}}$ & $0.001_{\pm \text{0.005}}$ & $0.001_{\pm \text{0.002}}$ &  $0.002_{\pm \text{0.003}}$ & $0.001_{\pm \text{0.001}}$&  $0.000_{\pm \text{0.005}}$ & $0.000_{\pm \text{0.001}}$\\
    & \textbf{Sim}$\boldsymbol{_{\cos}}(\downarrow)$ & $0.800_{\pm \text{0.006}}$ & $0.945_{\pm \text{0.004}}$ & $0.964_{\pm \text{0.006}}$ & $0.943_{\pm \text{0.005}}$ & $0.966_{\pm \text{0.001}}$ &  $0.978_{\pm \text{0.001}}$ & $0.981_{\pm \text{0.001}}$\\
    & \textbf{\#Pert.}$(\downarrow)$ & $15_{\pm \text{0.000}}$ & $15_{\pm \text{0.000}}$ & $15_{\pm \text{0.000}}$ & $15_{\pm \text{0.000}}$& $15_{\pm \text{0.000}}$  &  $15_{\pm \text{0.000}}$ & $15_{\pm \text{0.000}}$ \\
    & \textbf{$\Delta$Label}$(\downarrow)$ &  $0.051_{\pm \text{0.012}}$ & $0.011_{\pm \text{0.007}}$ & $0.012_{\pm \text{0.004}}$ & $0.009_{\pm \text{0.006}}$ & $0.004_{\pm \text{0.002}}$ &  $0.002_{\pm \text{0.002}}$ & $0.004_{\pm \text{0.002}}$\\
    & \textbf{$\Delta$Prob}$(\downarrow)$ & $0.048_{\pm \text{0.003}}$ & $0.009_{\pm \text{0.005}}$ & $0.006_{\pm \text{0.001}}$ & $0.008_{\pm \text{0.001}}$ & $0.006_{\pm \text{0.001}}$&  $0.003_{\pm \text{0.001}}$ & $0.003_{\pm \text{0.001}}$\\
    \hline
    \multirow[c]{6}{*}{\rotatebox{90}{Rnd. Rewiring}} & \textbf{P. GEA}$(\downarrow)$ & $0.149_{\pm \text{0.012}}$ & $0.045_{\pm \text{0.005}}$ & $0.069_{\pm \text{0.004}}$ & $0.067_{\pm \text{0.004}}$ & $0.020_{\pm \text{0.002}}$ &  $0.013_{\pm \text{0.001}}$ & $0.020_{\pm \text{0.002}}$\\
    & \textbf{$\Delta$GEA}$(\uparrow)$ & $0.530_{\pm \text{0.043}}$ & $0.596_{\pm \text{0.022}}$ & $0.425_{\pm \text{0.010}}$ & $0.402_{\pm \text{0.003}}$ & $0.419_{\pm \text{0.014}}$ &  $0.293_{\pm \text{0.025}}$ & $0.409_{\pm \text{0.012}}$\\
    & \textbf{Sim}$\boldsymbol{_{\cos}}(\downarrow)$ & $0.723_{\pm \text{0.031}}$ & $0.941_{\pm \text{0.006}}$ & $0.975_{\pm \text{0.006}}$ & $0.959_{\pm \text{0.008}}$ & $0.951_{\pm \text{0.004}}$ &  $0.970_{\pm \text{0.002}}$ & $0.967_{\pm \text{0.009}}$\\
    & \textbf{\#Pert.}$(\downarrow)$ & $16_{\pm \text{0.000}}$ & $16_{\pm \text{0.000}}$ & $16_{\pm \text{0.000}}$ & $16_{\pm \text{0.000}}$ & $16_{\pm \text{0.000}}$ &  $16_{\pm \text{0.000}}$ & $16_{\pm \text{0.000}}$\\
    & \textbf{$\Delta$Label}$(\downarrow)$ &  $0.274_{\pm \text{0.030}}$ &  $0.199_{\pm \text{0.022}}$ &  $0.202_{\pm \text{0.004}}$&  $0.184_{\pm \text{0.011}}$ &  $0.214_{\pm \text{0.007}}$ &  $0.159_{\pm \text{0.013}}$ &  $0.274_{\pm \text{0.014}}$\\
    & \textbf{$\Delta$Prob}$(\downarrow)$ & $0.234_{\pm \text{0.016}}$ & $0.210_{\pm \text{0.013}}$ & $0.198_{\pm \text{0.007}}$ & $0.200_{\pm \text{0.003}}$ & $0.221_{\pm \text{0.005}}$ &  $0.174_{\pm \text{0.013}}$ & $0.259_{\pm \text{0.005}}$\\
    \bottomrule
\end{tabular}
}
	\label{tab:algo_summary}
\end{table*}

From Table~\ref{tab:algo_summary}, we have the following main observations:

\textbf{Observation 1.} \textit{The larger the subgraph size, the lower the original explanation accuracy of PGExplainer and thus the poorer the attack performance} (i.e., the smaller the value of $\Delta \text{GEA}$). For this, compare the results on Syn2 (Subgraph Size = 6) to those on Syn3 (Subgraph Size = 10).

\textbf{Observation 2.} \textit{The type of motifs (namely ground-truth explanations) appears to have minimal impact on both explanation accuracy and attack performance.}  This can be obtained by  comparing the results on Syn3 (Motif =``House") with those on Syn4 (Motif =``Circle").

\textbf{Observation 3.} \textit{The larger the connection probability (namely node degree), the lower the original explanation accuracy of PGExplainer and thus the poorer the attack performance} (i.e., the smaller the value of $\Delta \text{GEA}$). This can be obtained by  comparing the results on Syn3 (P(Connection) = 0.06), Syn5 (P(Connection) = 0.12), and Syn6 (P(Connection) = 0.20).

\textbf{Observation 4.} \textit{The more subgraphs, the lower the original explanation accuracy of PGExplainer and thus the poorer the attack performance} (i.e., the smaller the value of $\Delta \text{GEA})$. This can be obtained by  comparing the results on Syn3 (\#Subgraphs=50) with those on Syn7 (\#Subgraphs=100).

\textbf{Answer to RQ1.} From Table~\ref{tab:algo_summary}, we can see that GXAttack can achieve a $\Delta \text{GEA}$ ranging from $0.088$ to $0.267$ while using a very small perturbation budget (less than 7.1 on all datasets) and keeping the predicted labels nearly unchanged (i.e., $\Delta \text{Label}$ is nearly zero) after perturbation.

\textbf{Answer to RQ2.} In contrast, the random flipping attack employs a budget of 15 but achieves $\Delta \text{GEA}$ of almost zero on all datasets, indicating it is ineffective. Meanwhile, the random rewiring attack uses a budget of 16 and achieves a very large $\Delta \text{GEA}$ (ranging from $0.293$ to $0.596$). However, this comes at the cost of large $\Delta \text{Label}$ (ranging from $0.159$ to $0.274$). In other words, the predicted labels of many nodes have been changed after the attacks, and this violates our objective.

\subsection{Experiments to Answer RQ3: Attack Transferability}
\label{subsec:RQ3}

\textbf{Answer to RQ3.} From 
Table~\ref{tab:transfer_attack}, we can see that some post-hoc explainers, including PGMExplainer, and IG,
suffer from poor explanation accuracy on the original graphs, even giving lower accuracy than the Random Explainer. As a result, investigating the attack transferability on these explainers may not be meaningful. On the contrary, like PGExplainer, explainers such as GradCAM, GNNExplainer, SubgraphX, and GNN-LRP can achieve good performance on most original graphs. For those, the attacks optimized for PGExplainer are also harmful, leading to a dramatic degradation of explanation accuracy. For instance, the explanation accuracy of GNNexplainer decreases from 0.711 to 0.397 on Syn2. This demonstrates the transferability of the attacks generated by GXAttack.

\subsection{Sensitivity Analysis}
\label{subsec:SenAna}
\begin{figure}[h]
    \centering
    \includegraphics[width=1\linewidth]{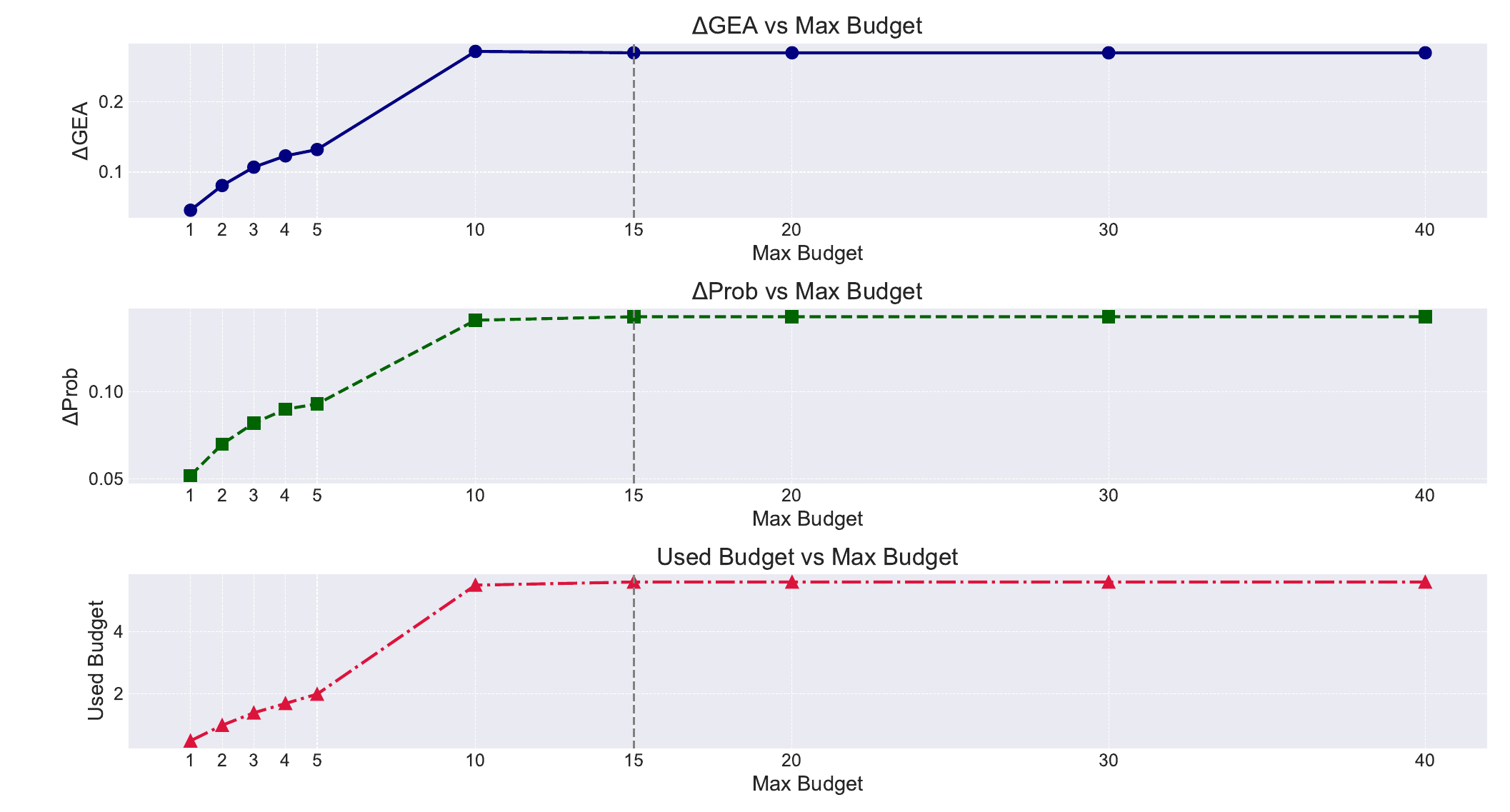}
    \caption{Sensitivity analysis w.r.t. maximally allowed perturbation budget (Max Budget) on Syn2 using GXAttack.
    \label{fig:Syn2_SA_Budget}}
\end{figure}

\begin{figure}[h]
    \centering
    \includegraphics[width=1\linewidth]{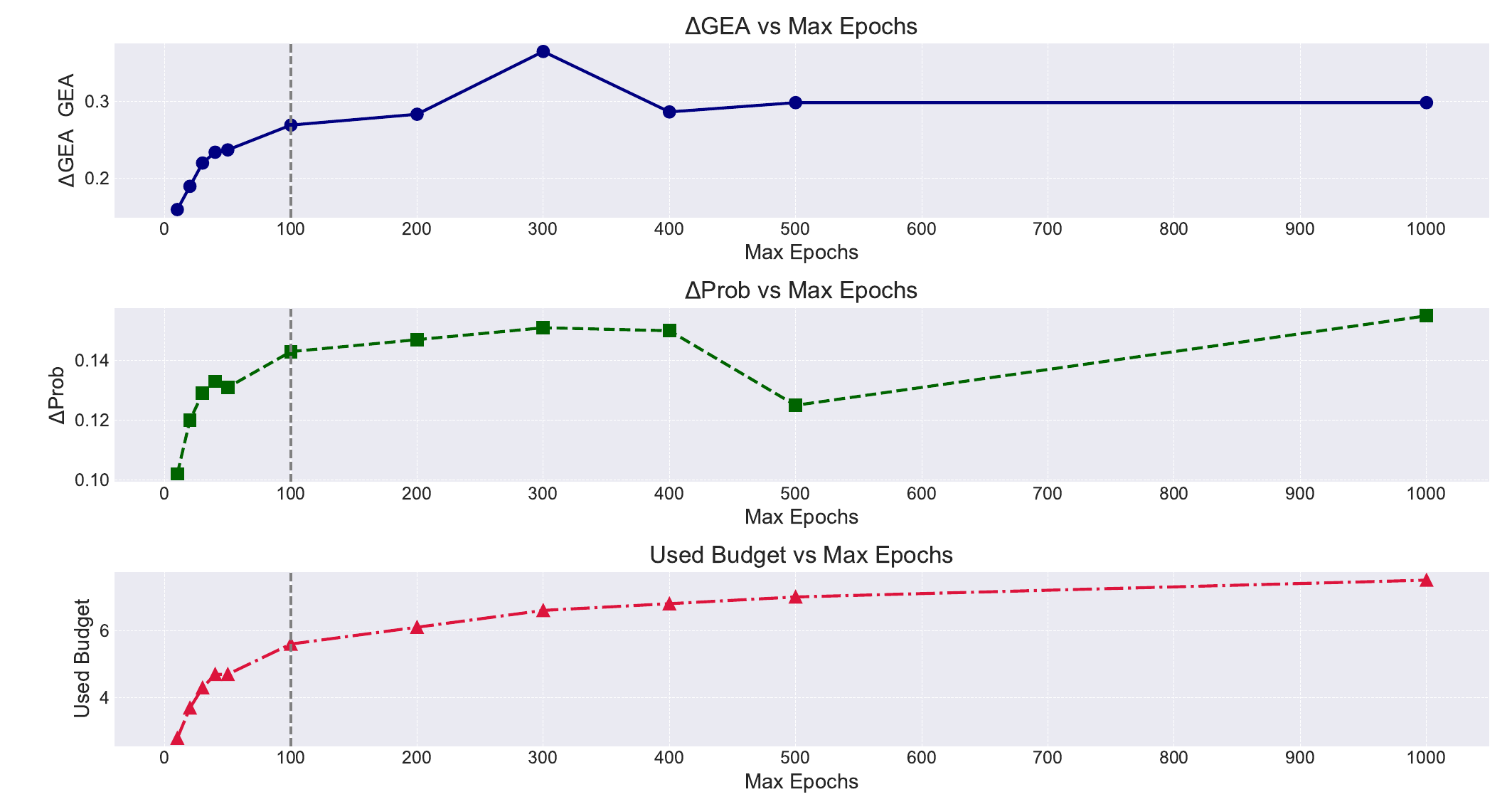}
    \caption{Sensitivity analysis w.r.t. maximal training epochs on Syn2 using GXAttack.}
    \label{fig:Syn2_SA_Epochs}
\end{figure}

Specifically, we perform analysis in terms of $\Delta$GEA, $\Delta$Prob and used budget when varying the following hyperparameters, respectively: 1) the loss trade-off hyperparameter; 2) the maximally allowed perturbation budget (Max Budget) in Figure~\ref{fig:Syn2_SA_Budget}; and 3) the maximal training epochs of attacker in Figure~\ref{fig:Syn2_SA_Epochs}.

\begin{table*}[h]
	\caption{Transferability of attacks on synthetic datasets Syn1-6 with a typical run (Results on Syn7 are omitted due to excessive computation time). We report the explanation accuracy (i.e., GEA) on the original graph $\rightarrow$ explanation accuracy on the perturbed graph.}
	\centering
\resizebox{\linewidth}{!}{
	\begin{tabular}{ccccccc}
		\toprule
		\textbf{Explainer} & Syn1 & Syn2 & Syn3 & Syn4 & Syn5 & Syn6 \\
		\midrule
        \textbf{PGExplainer} & 0.728 $\rightarrow$ 0.510 & 0.648 $\rightarrow$ 0.347 & 0.485 $\rightarrow$ 0.304 & 0.471 $\rightarrow$ 0.311 & 0.450 $\rightarrow$ 0.285 & 0.332 $\rightarrow$ 0.225 \\
         \hline 
        \textbf{GradCAM} & 0.745 $\rightarrow$ 0.551 & 0.708 $\rightarrow$ 0.397 & 0.510 $\rightarrow$ 0.300 & 0.511 $\rightarrow$ 0.310 & 0.455 $\rightarrow$ 0.278 & 0.414 $\rightarrow$ 0.241 \\
        \textbf{IG} & 0.118 $\rightarrow$ 0.104 & 0.134 $\rightarrow$ 0.103 & 0.136 $\rightarrow$ 0.101 & 0.109 $\rightarrow$ 0.092 & 0.116 $\rightarrow$ 0.085 & 0.078 $\rightarrow$ 0.065 \\
        \hline
        \textbf{GNNExplainer} & 0.738 $\rightarrow$ 0.550 & 0.711 $\rightarrow$ 0.397 & 0.539 $\rightarrow$ 0.341 & 0.514 $\rightarrow$ 0.327 & 0.483 $\rightarrow$ 0.283 & 0.431 $\rightarrow$ 0.205 \\
        \textbf{SubgraphX} & 0.113 $\rightarrow$ 0.015 & 0.532 $\rightarrow$ 0.375 & 0.296 $\rightarrow$ 0.141 & 0.417 $\rightarrow$ 0.282 & 0.394 $\rightarrow$ 0.312 & 0.408 $\rightarrow$ 0.305 \\
        \hline
        \textbf{PGMExplainer} & 0.015 $\rightarrow$ 0.064 & 0.046 $\rightarrow$ 0.039 & 0.031 $\rightarrow$ 0.034 & 0.036 $\rightarrow$ 0.041 & 0.038 $\rightarrow$ 0.043 & 0.037 $\rightarrow$ 0.040 \\
        \hline
        \textbf{GNN-LRP} & 0.752 $\rightarrow$ 0.573 &	0.713 $\rightarrow$ 0.394 & 0.509 $\rightarrow$ 0.300 &0.515 $\rightarrow$ 0.311 & 0.457 $\rightarrow$ 0.284 & 0.410 $\rightarrow$ 0.241 \\
        \hline
        \textbf{RandomExplainer} & 0.184 $\rightarrow$ 0.146 & 0.170 $\rightarrow$ 0.096 & 0.135 $\rightarrow$ 0.102 & 0.124 $\rightarrow$ 0.090 & 0.128 $\rightarrow$ 0.100 & 0.133 $\rightarrow$ 0.086 \\
        \bottomrule
	\end{tabular}}
	\label{tab:transfer_attack}
\end{table*}

\textbf{Trade-off Hyperparameter vs Attack Performance.} Our sensitivity analysis shows that the attack performance (in term of $\Delta$GEA), $\Delta$Prob, and the used budget are nearly constant when we vary the trade-off hyper-parameter $\beta$'s value from $0.000001$ to 10. We omit the figures here.

\textbf{Attack Budget vs Attack Performance.} From Figure~\ref{fig:Syn2_SA_Budget}, we can see that as the Max Budget increases from 1 to 10, $\Delta$GEA and $\Delta$Prob rise sharply, showing large gains from higher allowed budgets. Beyond a (maximally  allowed) budget of 10, $\Delta$GEA and $\Delta$Prob both stabilize, indicating no further gains. The underlying reason is that the used budget stop increasing after 10 (maximally allowed) when we set the training epochs as 100. In Tables~\ref{tab:algo_summary} and \ref{tab:transfer_attack}, we report the results with a maximum budget equal to 15.

\textbf{Attack Training Epochs vs Attack Performance.} From Figure~\ref{fig:Syn2_SA_Epochs}, we observe a steady increase in the used budget as epochs increase. However  $\Delta$Prob first increases and then decreases (due to overfitting) when increasing the training epochs. Similarly, $\Delta$GEA first increases and then decreases (due to overfitting) with the increase of training epochs. For computation time reason, we only report the results with 100 epochs in Tables~\ref{tab:algo_summary} and \ref{tab:transfer_attack}. However, it is important to note that the attack performance (in terms of $\Delta$GEA) can be  largely improved if we increase the number of training epochs. For instance, $\Delta$GEA can be improved from 0.269 to 0.364 if we increase the training epochs from 100 to 300.

\subsection{Property Analysis}
\label{subsec:ProAna}
We also conducted some property analysis of GXAttack in terms of the prediction confidence, and node degree.

\begin{figure*}
    \centering
    \begin{minipage}{0.3\textwidth}
        \centering
        \includegraphics[width=\linewidth]{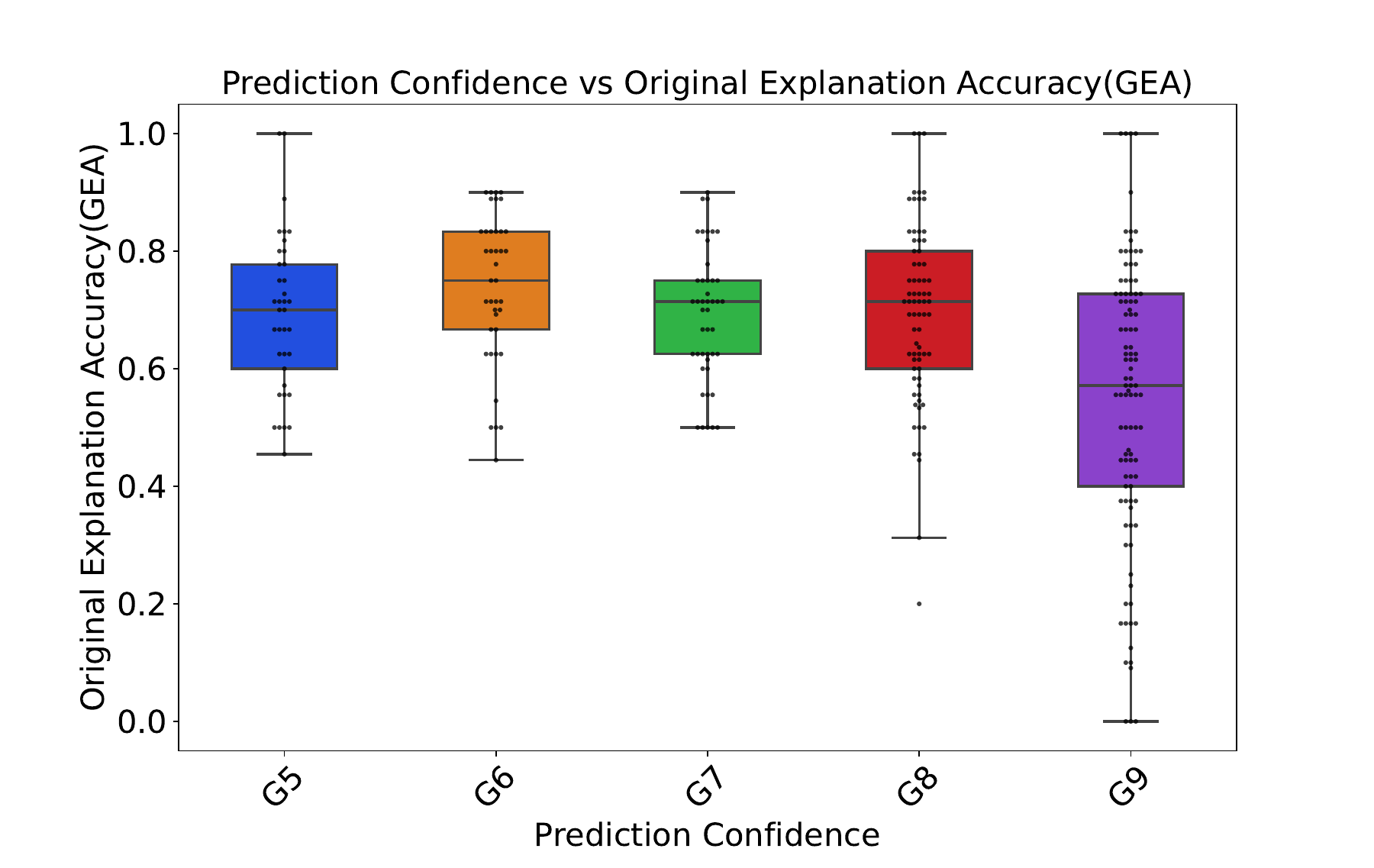}
        \caption*{(a) Prediction confidence vs. original explanation accuracy on Syn2.}
    \end{minipage}%
    \hfill
    \begin{minipage}{0.32\textwidth}
        \centering
        \includegraphics[width=\linewidth]{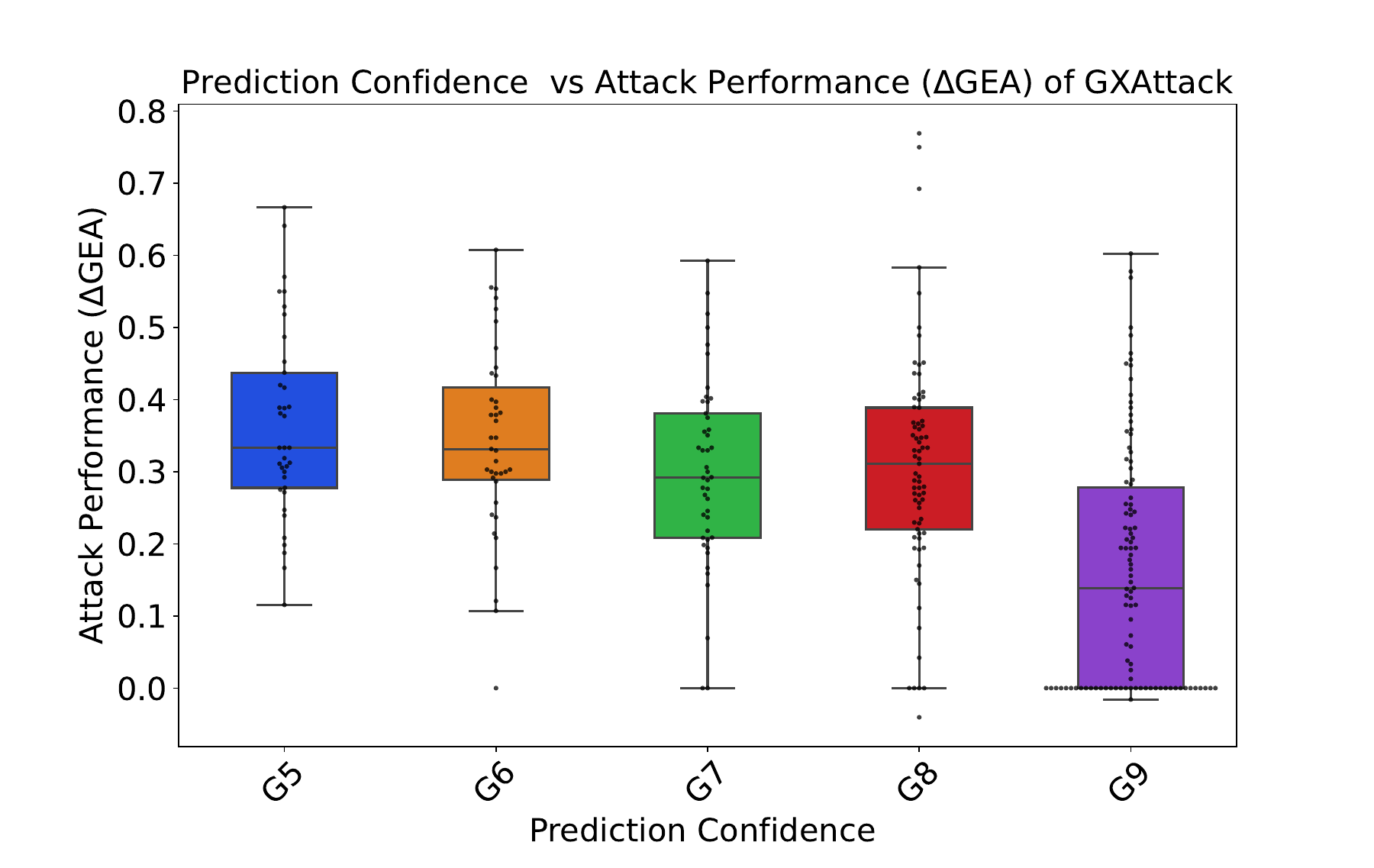}
        \caption*{(b) Prediction confidence vs. attack performance ($\Delta\text{GEA}$) on Syn2.}
    \end{minipage}%
    \hfill
    \begin{minipage}{0.32\textwidth}
        \centering
        \includegraphics[width=\linewidth]{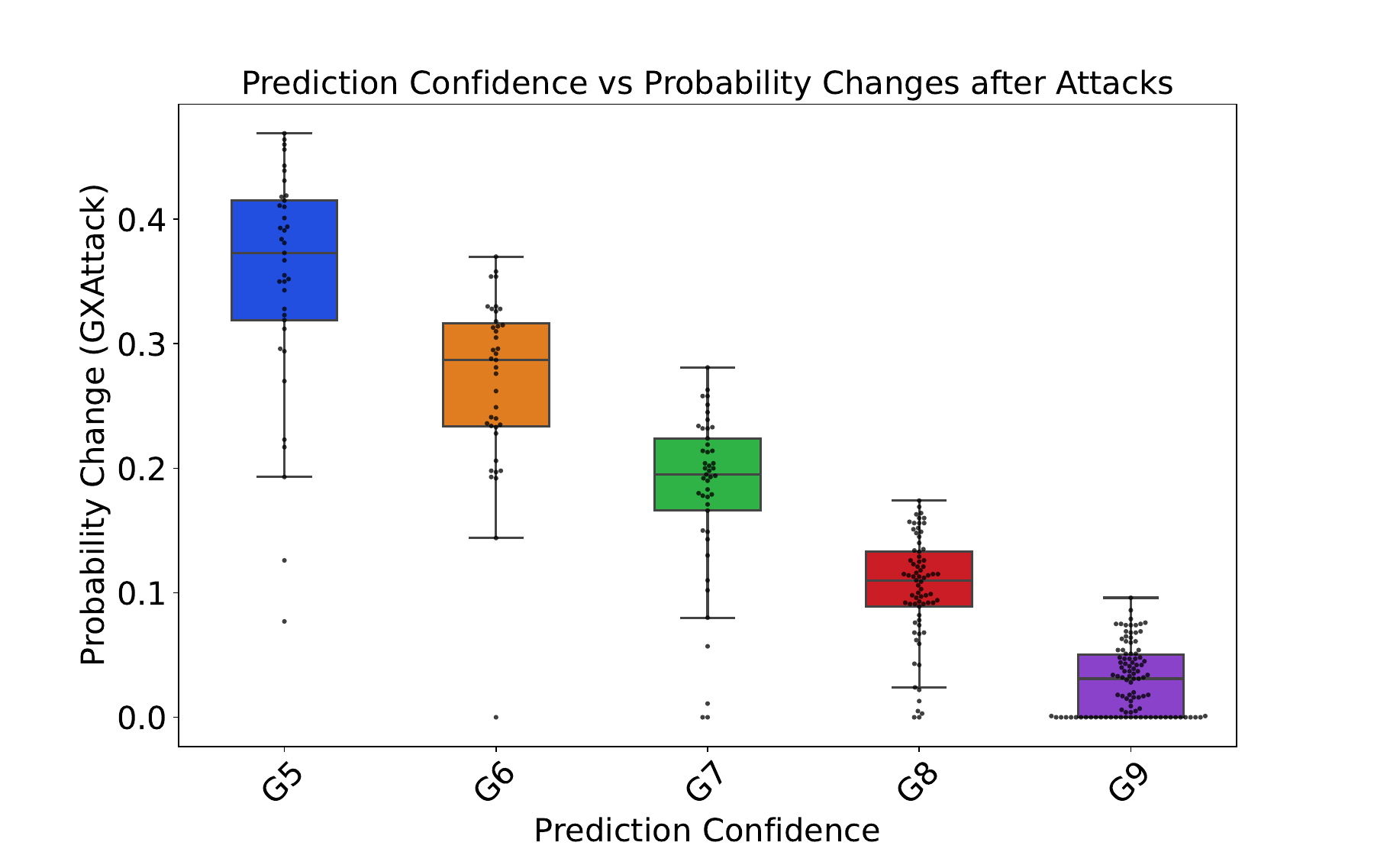}
        \caption*{(c) Prediction confidence vs. probability change after attacks on Syn2.}
    \end{minipage}
    
    \caption{Property analysis regarding prediction confidence on Syn2. Other datasets show consistent results and are omitted.}
    \label{fig:PreConProAnalysis}
\end{figure*}

\textbf{Property Analysis on Prediction Confidence.} From Figure~\ref{fig:PreConProAnalysis}(a), we can clearly see that nodes in `G9' tend to have lower original explanation accuracy. As a result, as shown in Figure~\ref{fig:PreConProAnalysis}(b), the attack performance of GXAttack on these nodes (in `G9') tend to be less successful (i.e., smaller $\Delta\text{GEA}$). This is because the nodes in `G9' leave no much room for perturbations under the constraint that the predictions remain unchanged, namely much smaller probability changes as shown in Figure~\ref{fig:PreConProAnalysis}(c).

\begin{figure*}[]
    \centering
    \begin{minipage}{0.32\textwidth}
        \centering
        \includegraphics[width=\linewidth]{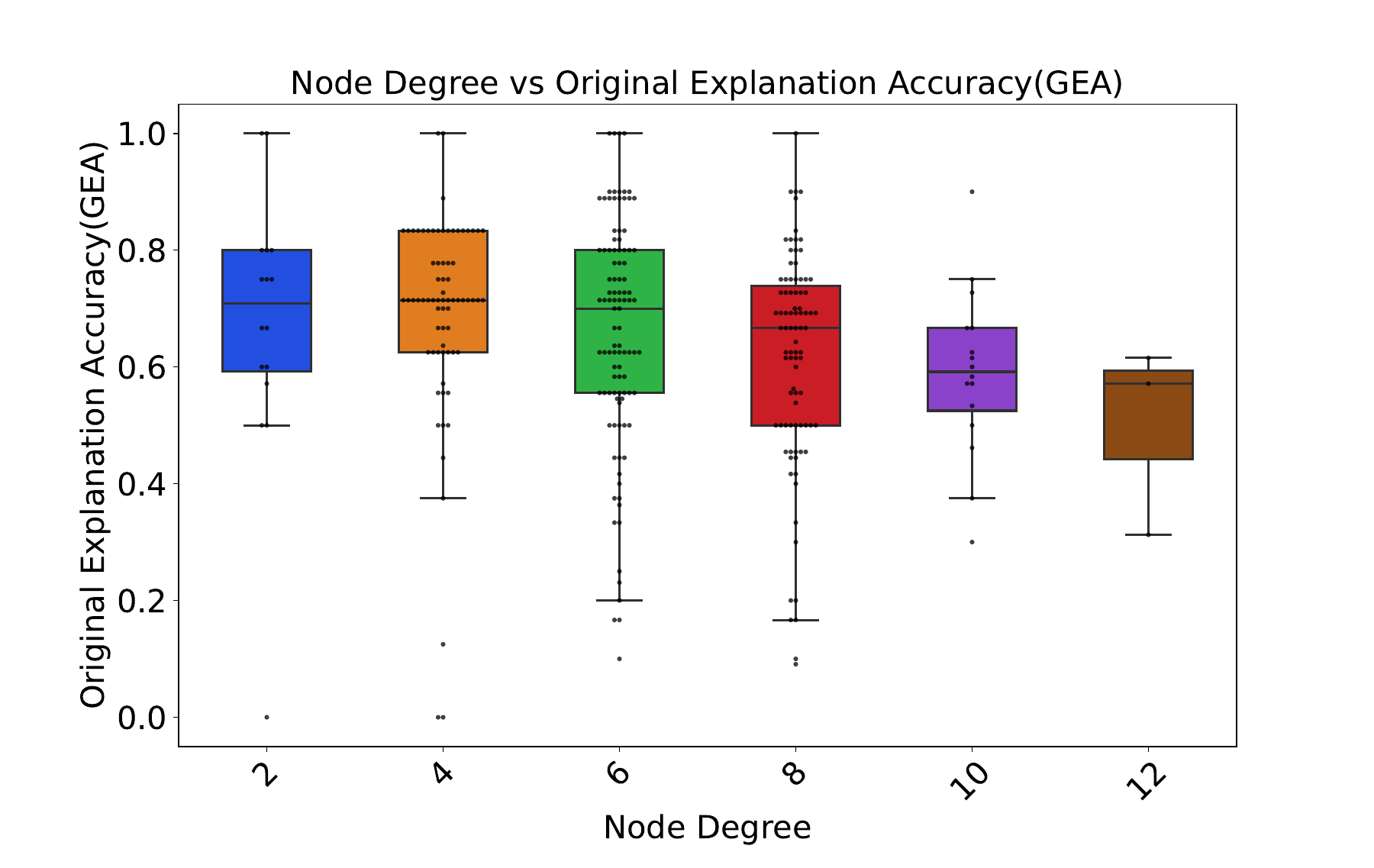}
        \caption*{(a) Node degree vs. original explanation accuracy on Syn2.}
        \label{fig:Syn2_NodeDegVsOGEA}
    \end{minipage}%
    \hfill
    \begin{minipage}{0.32\textwidth}
        \centering
        \includegraphics[width=\linewidth]{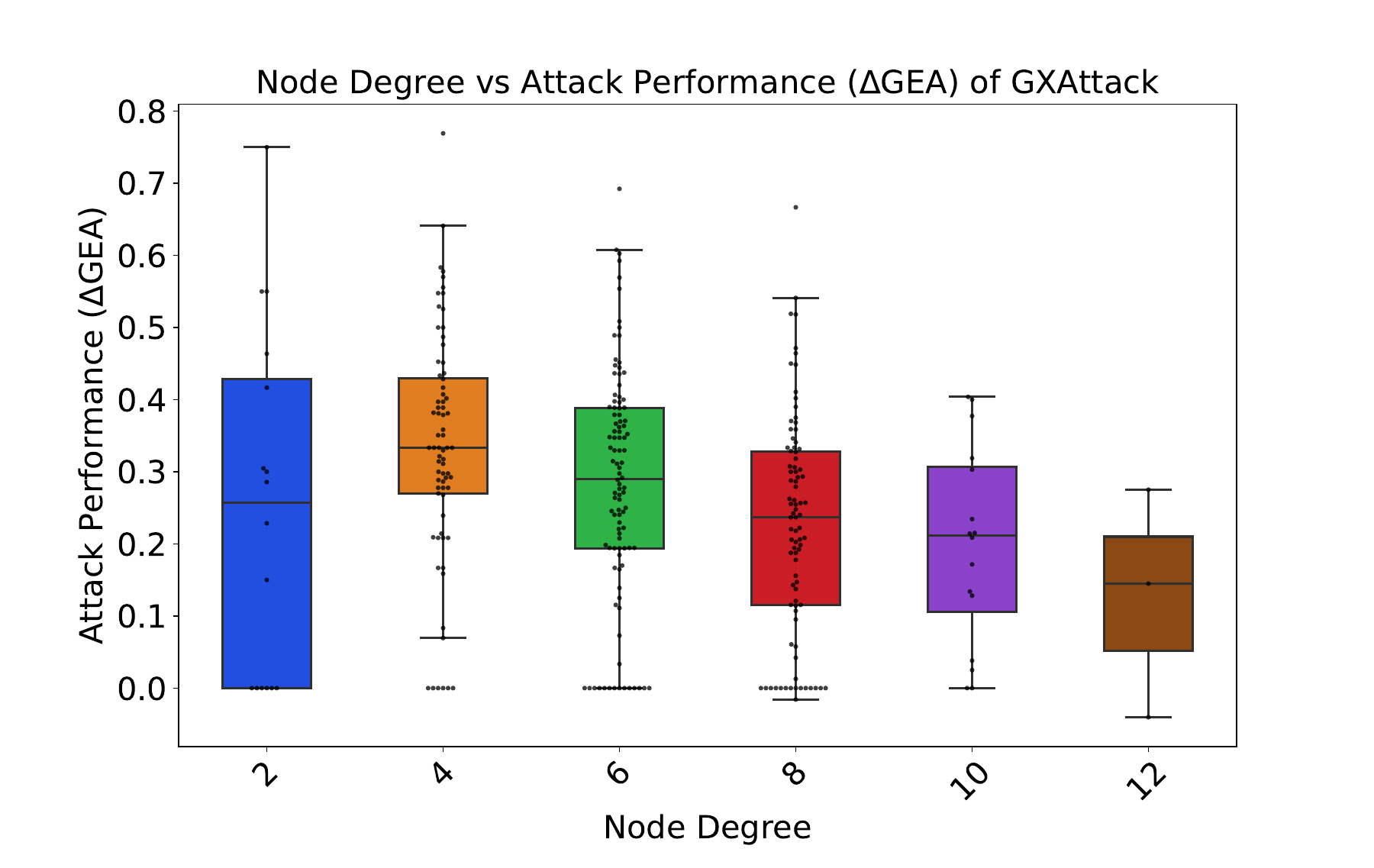}
        \caption*{(b) Node degree vs. attack performance ($\Delta\text{GEA}$) on Syn2.}
        \label{fig:Syn2_NodeDegVsDeltaGEA}
    \end{minipage}%
    \hfill
    \begin{minipage}{0.32\textwidth}
        \centering
        \includegraphics[width=\linewidth]{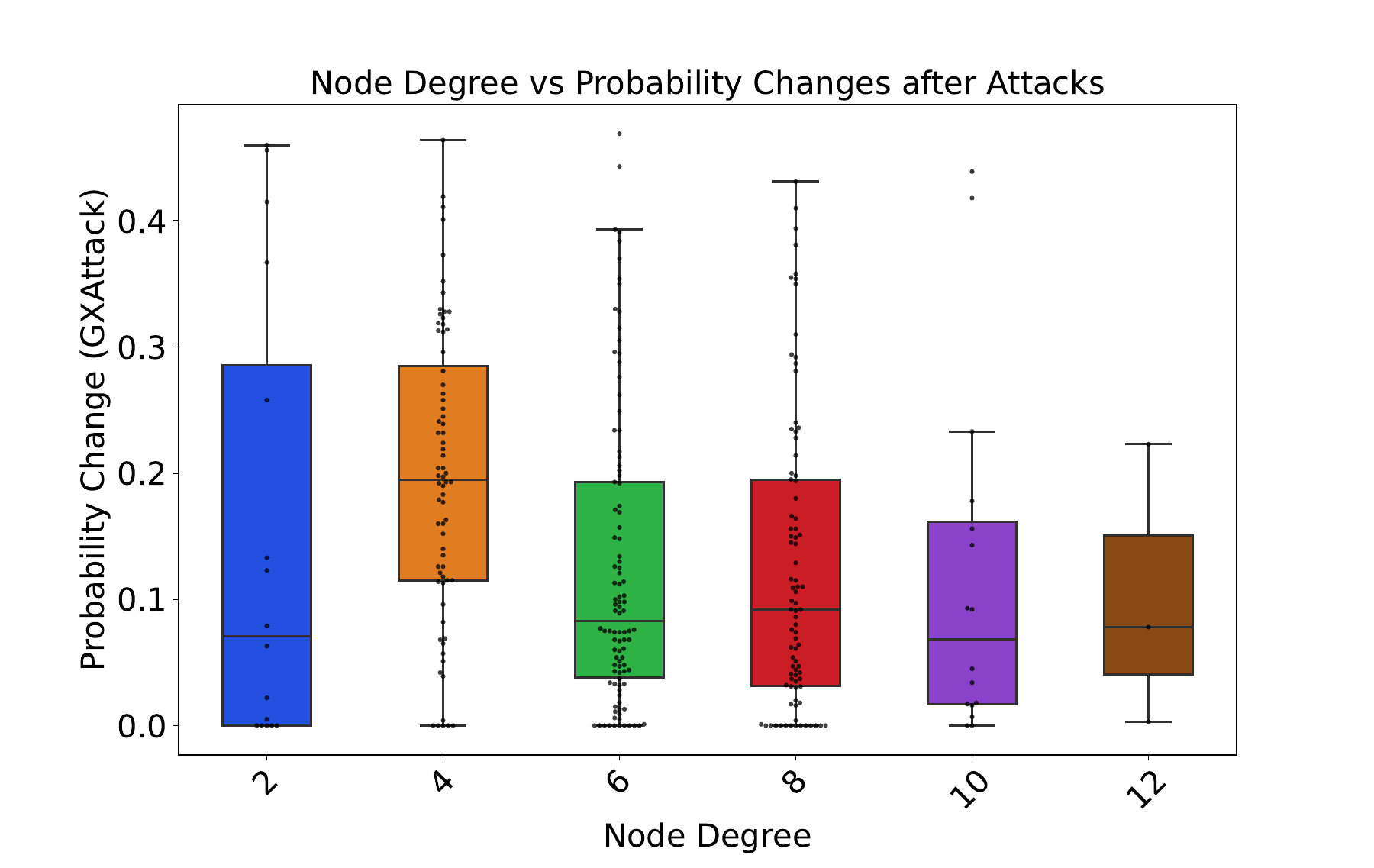}
        \caption*{(c) Node degree vs. probability change after attacks on Syn2.}
        \label{fig:Syn2_NodeDegVsProbCh}
    \end{minipage}
    \caption{Property analysis regarding node degree on Syn2. Other datasets show consistent results and thus are omitted.}
    \label{fig:NodeDegreeProAnalysis}
\end{figure*}

\textbf{Property Analysis on Node Degree.} From Figure~\ref{fig:NodeDegreeProAnalysis}(a), we can see that nodes with higher degrees tend to have lower original explanation accuracy. Figure~\ref{fig:NodeDegreeProAnalysis}(b) shows that the attack performance of GXAttack on these nodes with higher degrees tends to be less successful (i.e., smaller $\Delta\text{GEA}$). Possible reasons are that 1) the original explanation accuracy tends to be lower; and 2) small perturbations on nodes with high degrees have limited impact on the predictions, resulting in smaller probability changes as shown in Figure~\ref{fig:NodeDegreeProAnalysis}(c).

\section{Conclusions}
\label{sec:conclusions}
GNN explanations for enhancing transparency and trust of graph neural networks are becoming increasingly important in decision-critical domains. Our research reveals that existing GNN explanation methods are vulnerable to adversarial perturbations that strongly change the explanations without affecting the predictions. This vulnerability undermines the reliability of these methods in crucial settings. Specifically, we introduced \textit{GXAttack}, an optimization-based adversarial method to 
assess the robustness of post-hoc GNN explanations. GXAttack's effectiveness underscores the need for new evaluation standards that incorporate adversarial robustness as a fundamental aspect. Future research should focus on developing explanation methods that are both interpretable and robust against adversarial attacks, as enhancing the stability of GNN explanations is crucial for ensuring their consistency and reliability in practical applications.

\section*{Acknowledgements}
\textbf{Zhong Li} and \textbf{Matthijs van Leeuwen}: this publication is part of the project Digital Twin with project number P18-03 of the research programme TTW Perspective, which is (partly) financed by the Dutch Research Council (NWO). \textbf{Simon Geisler}: This research was supported by the Helmholtz Association under the joint research school ``Munich
School for Data Science – MUDS".

\bibliographystyle{IEEEtran}
\bibliography{references.bib}


\appendix

\section*{Appendix A: Projected Gradient Descent}
\label{app:PGD}

After each gradient update, $\Tilde{\mathbf{P}}_{t} \in \mathbb{R}^{n\times n}$, although it must lie in $[0,1]^{n\times n}$ for its interpretation as a Bernoulli random variable. Moreover, we must ensure that sampling $\mathbf{P}_{t} \in \{0,1\}^{n\times n} \sim \text{Bernoulli}(\Tilde{\mathbf{P}}_{t})$ will yield a discrete perturbation $\hat{\mathbf{A}} = \mathbf{A}\oplus\mathbf{P}_{t}$ obeying the admissible perturbations $\Vert \mathbf{A} - \hat{\mathbf{A}} \Vert_{0} < B$ with sufficient likelihood. Following \cite{xu2019topology,geisler2021robustness}, we ensure that $\mathbb{E}_{\mathbf{P}\sim\text{Bernoulli}(\Tilde{\mathbf{P}_t})}[\Vert\mathbf{P}\Vert_0] \leq B$ and then obtain the final perturbation form a small set of samples. To guarantee that $\Tilde{\mathbf{P}}_{t}$ parametrizes a valid distribution and that we obey the budget in expectation, we employ the projection $\prod_{\mathbb{E}[\text{Bernoulli}(\Tilde{\mathbf{P}}_{t})\leq B]}(\Tilde{\mathbf{P}}_{t})$.

The space spanned by $\Tilde{\mathbf{P}}_{t} \in [0,1]^{n\times n}$ describes a $n\times n$ dimensional hypercube intersected with the region below the $B$-simplex defining $\mathbb{E}[\text{Bernoulli}(\Tilde{\mathbf{P}}_{t}) = B]$. The region below the $B$-simplex can be expressed as $\vec{1}^\top \Tilde{\mathbf{P}} \vec{1} = \sum_{i,j} \Tilde{\mathbf{P}}_{i,j} \le B$. Thus, after each gradient update, we solve
\begin{align}
    &\prod_{\mathbb{E}[\text{Bernoulli}(\Tilde{\mathbf{P}})\leq B]}(\Tilde{\mathbf{P}}) = \arg\min_{\Tilde{\mathbf{S}}} \|\Tilde{\mathbf{P}} - \Tilde{\mathbf{S}}\|_F^2 \\
    & \qquad \text{subject to } \; 
    \sum_{i,j} \Tilde{S}_{i,j} \le B ~\text{and}~ \Tilde{\mathbf{S}} \in [0, 1]^{n \times n} \nonumber
\end{align}
for clipping to the hyperplane and projecting back towards the simplex using a bisection search. See \cite{xu2019topology} for details and the derivation.

\section*{Appendix B: Presumptions Checking Results}
\label{app:PreCheckRes}

\begin{figure}[]
    \centering
    
    \begin{minipage}{0.24\textwidth}
        \centering
        \includegraphics[width=\linewidth]{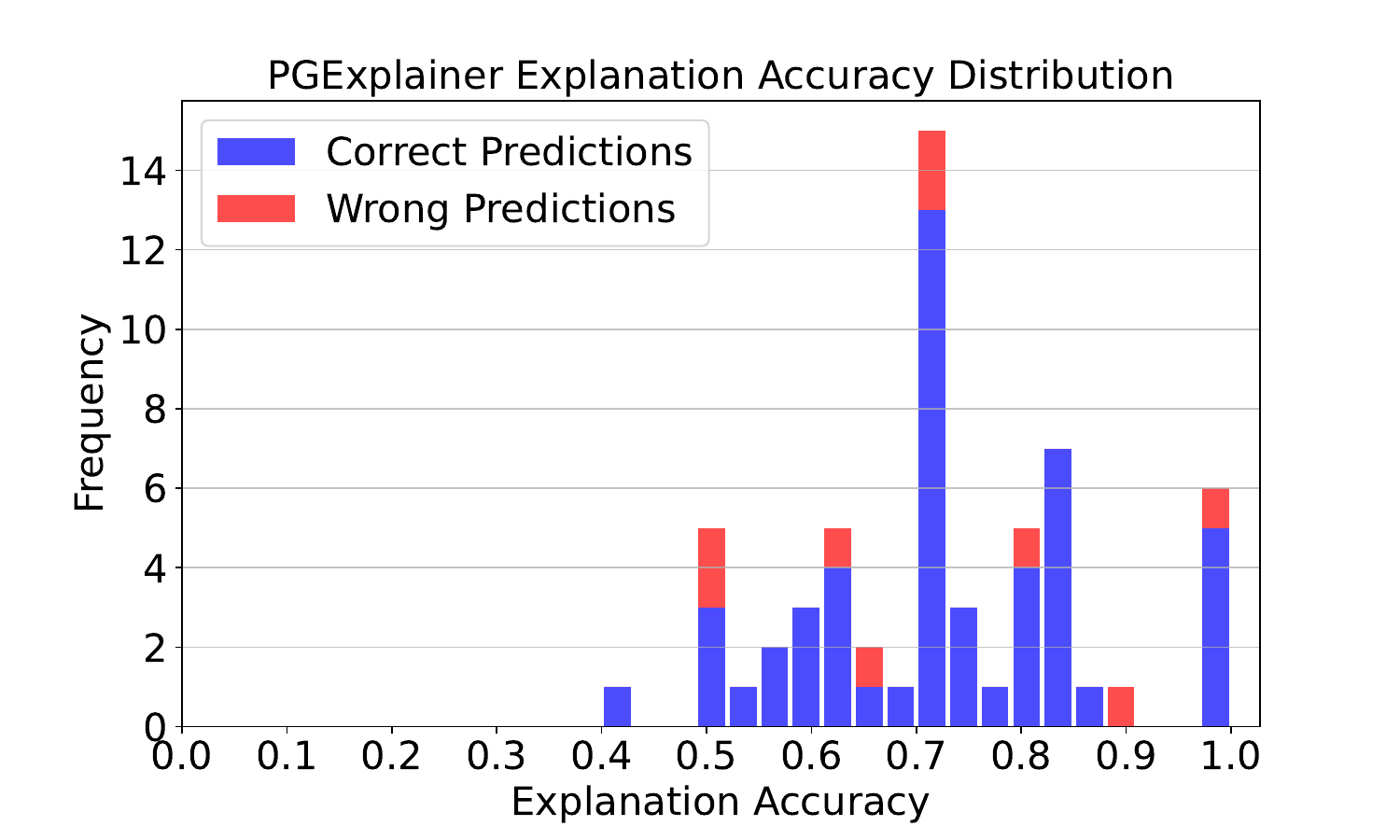}
        \caption*{(a) Syn1}
        \label{fig:ExpAccVSPreAcc_Syn1}
    \end{minipage}%
    \hfill
    \begin{minipage}{0.24\textwidth}
        \centering
        \includegraphics[width=\linewidth]{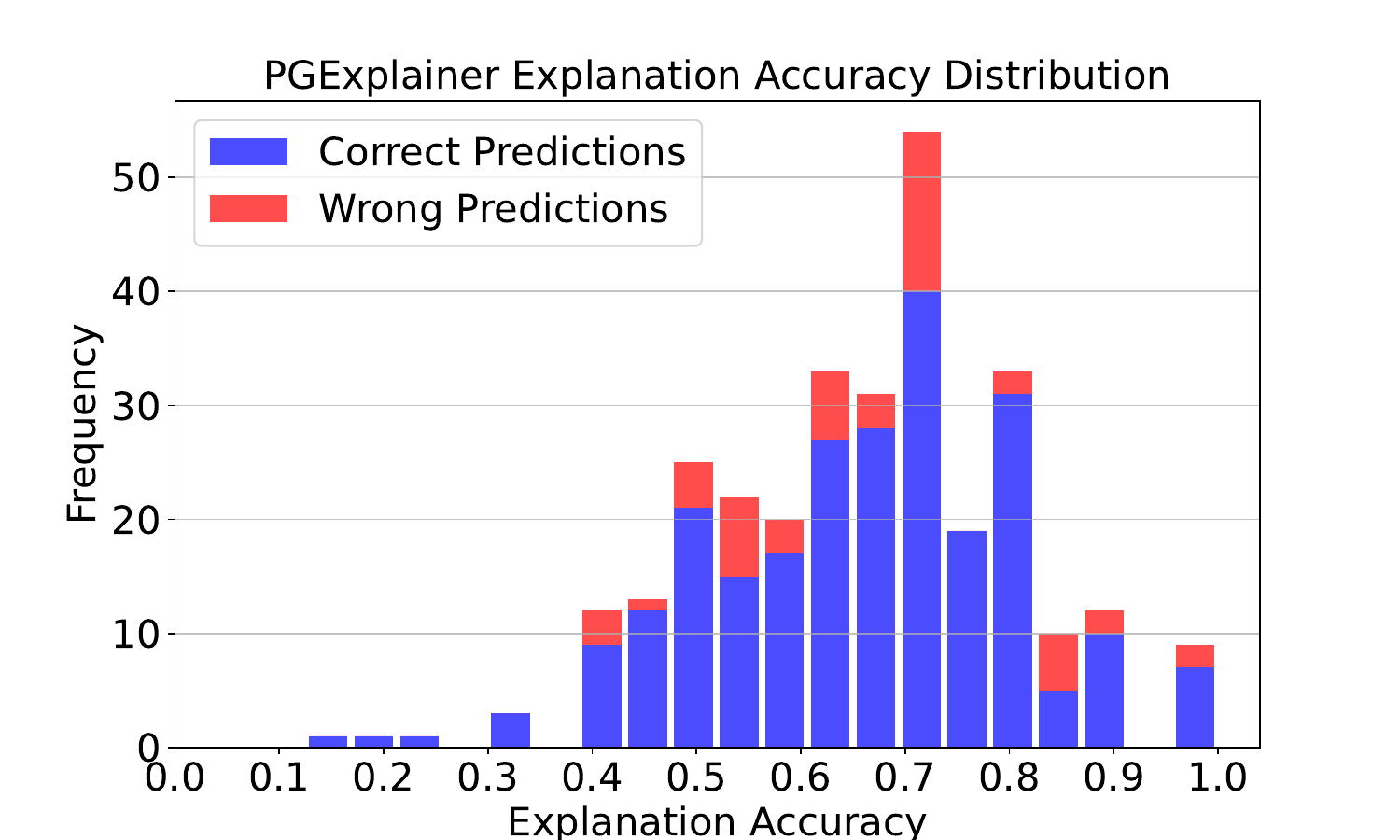}
        \caption*{(b) Syn2}
        \label{fig:ExpAccVSPreAcc_Syn2}
    \end{minipage}%
    \hfill
    \begin{minipage}{0.24\textwidth}
        \centering
        \includegraphics[width=\linewidth]{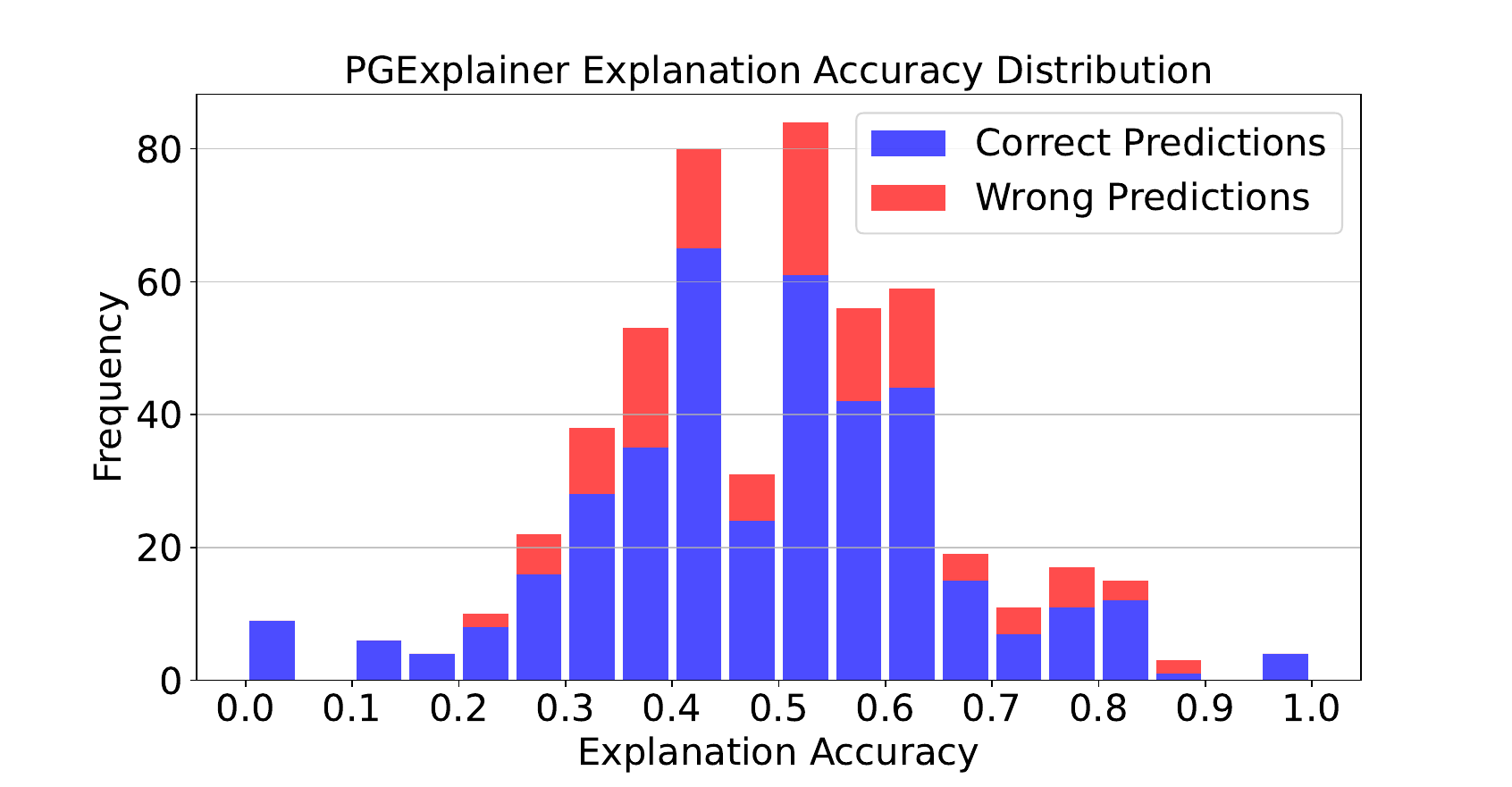}
        \caption*{(c) Syn3}
        \label{fig:ExpAccVSPreAcc_Syn3}
    \end{minipage}

    \vspace{0.5cm}  
    \begin{minipage}{0.24\textwidth}
        \centering
        \includegraphics[width=\linewidth]{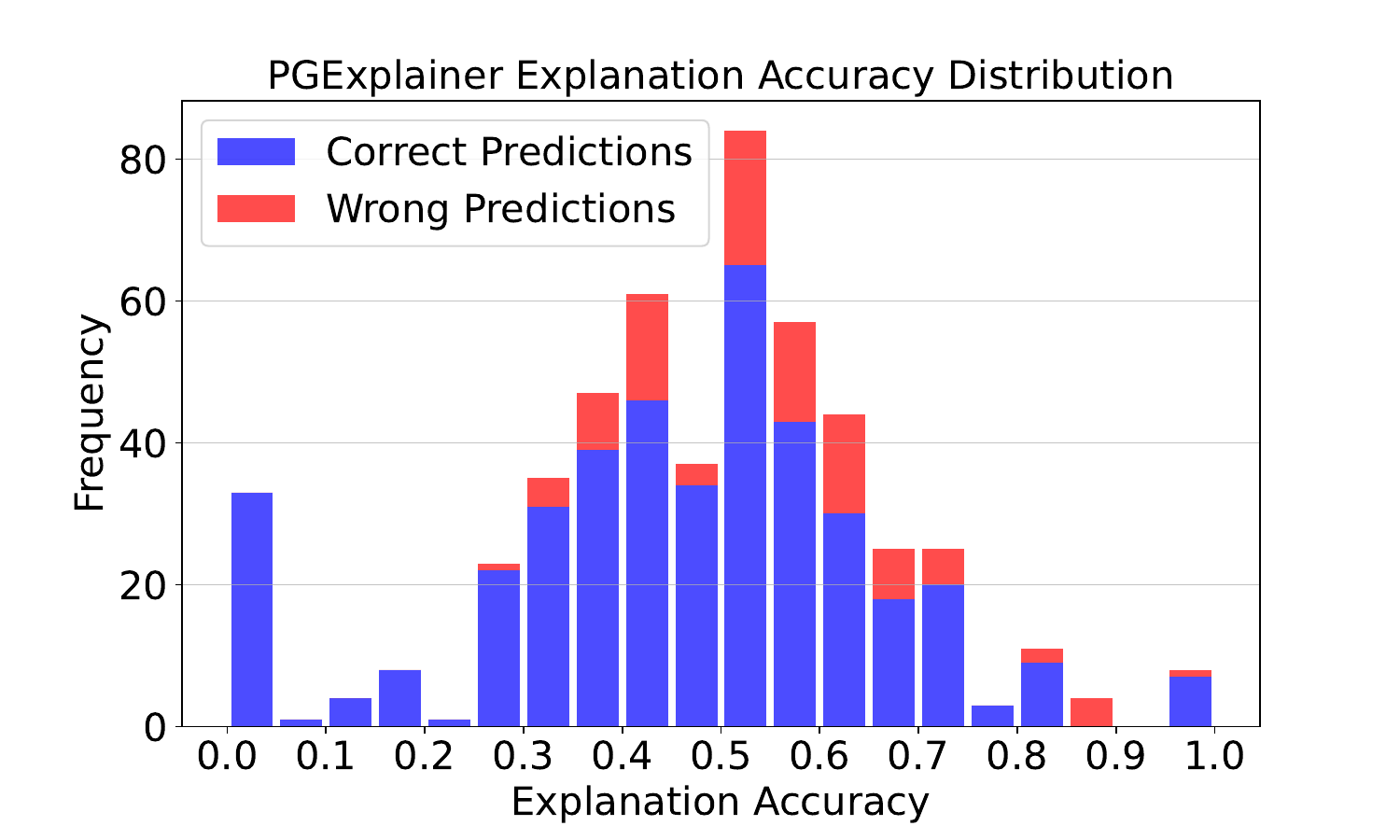}
        \caption*{(d) Syn4}
        \label{fig:ExpAccVSPreAcc_Syn4}
    \end{minipage}%
    \hfill
    \begin{minipage}{0.24\textwidth}
        \centering
        \includegraphics[width=\linewidth]{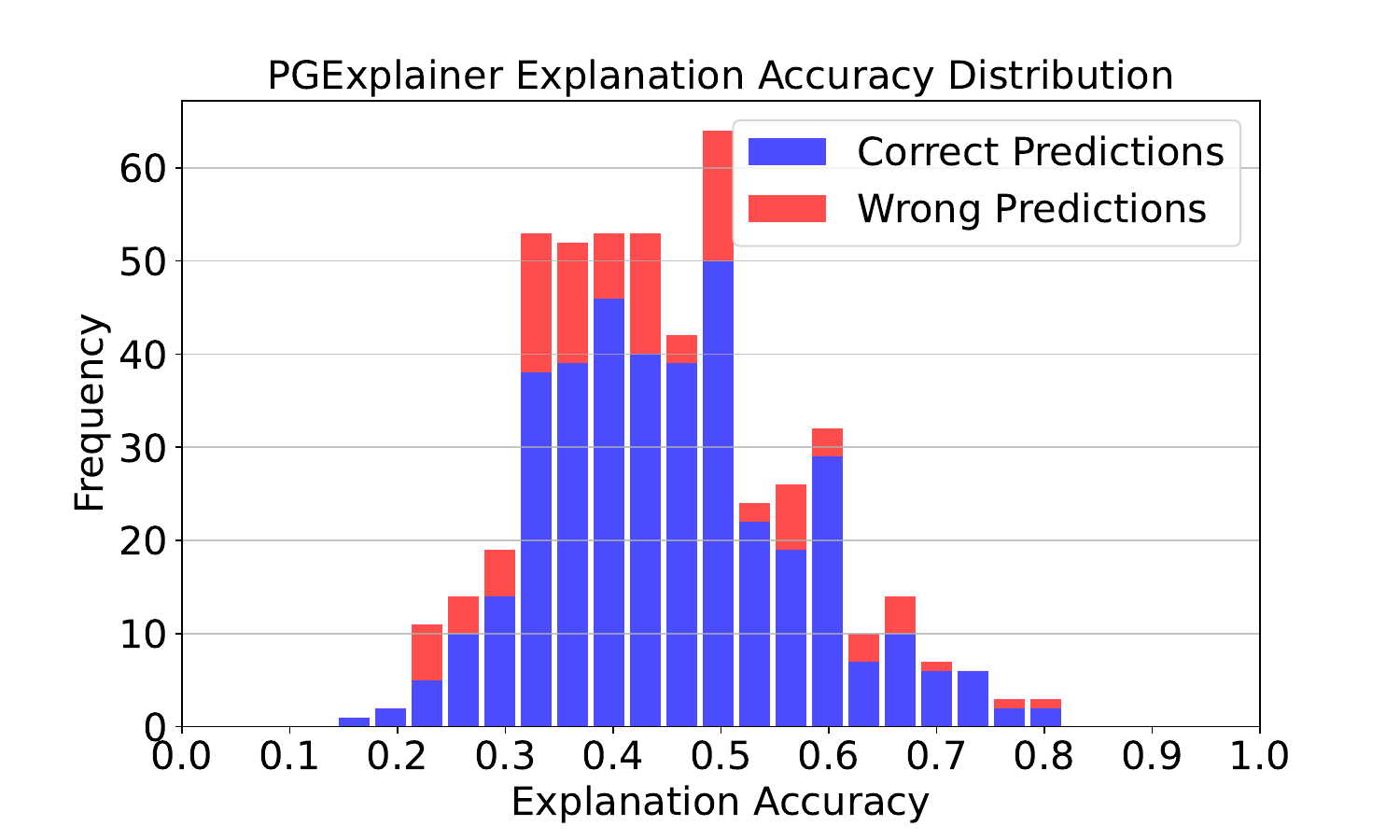}
        \caption*{(e) Syn5}
        \label{fig:ExpAccVSPreAcc_Syn5}
    \end{minipage}
    \begin{minipage}{0.24\textwidth}
        \centering
        \includegraphics[width=\linewidth]{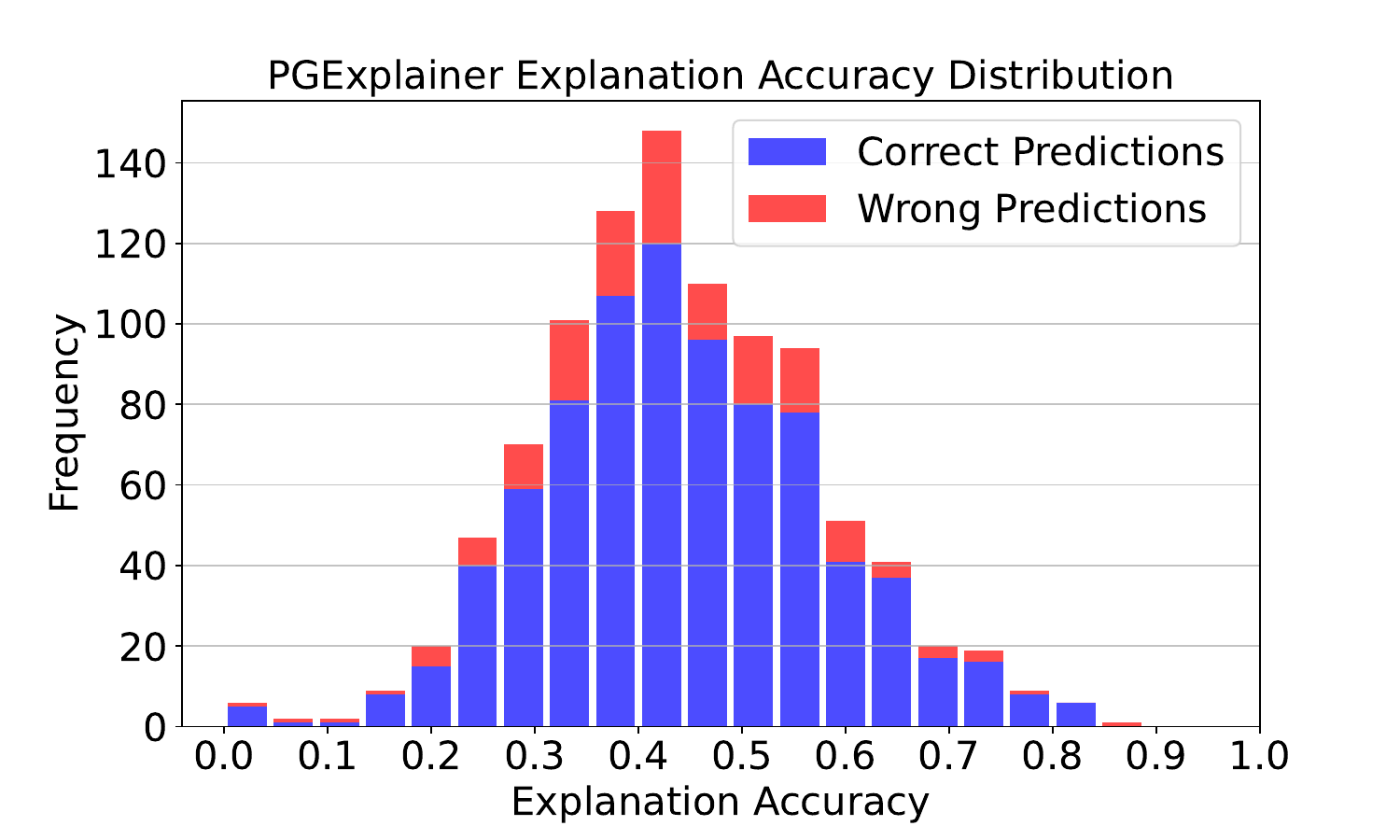}
        \caption*{(f) Syn6}
        \label{fig:ExpAccVSPreAcc_Syn6}
    \end{minipage}%
    \hfill
    \begin{minipage}{0.24\textwidth}
        \centering
        \includegraphics[width=\linewidth]{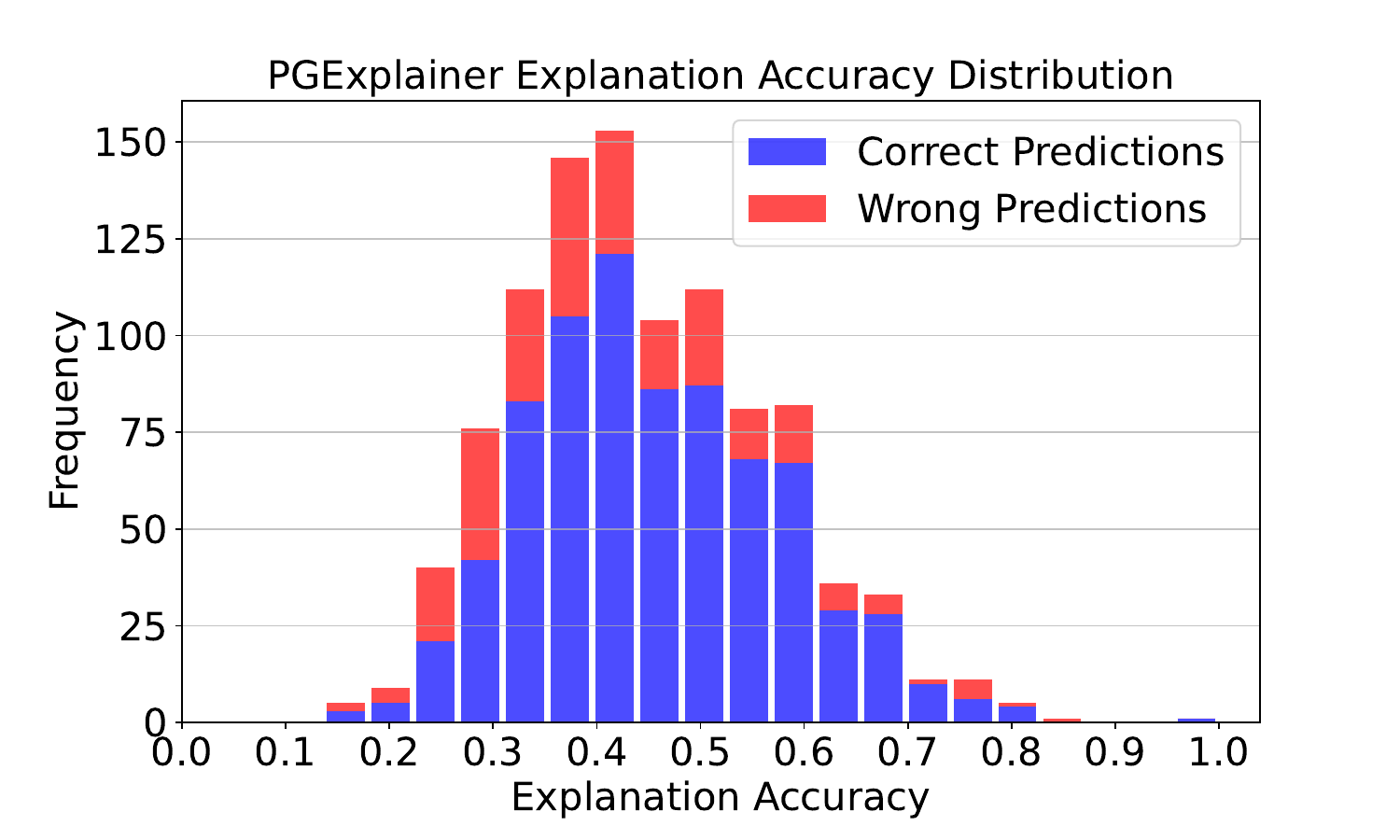}
        \caption*{(g) Syn7}
        \label{fig:ExpAccVSPreAcc_Syn7}
    \end{minipage}
    
    \caption{Comparative analysis of explanation accuracy with respect to prediction correctness on all synthetic datasets. The results show that explanation accuracy is not necessarily correlated with the prediction correctness. Particularly, we did \textbf{not} observe that wrongly predicted nodes tend to have lower explanation accuracy.}
    \label{fig:ExpAccVSPreAcc_All}
\end{figure}

\begin{table}[h]
	\caption{Complete results on Syn1-Syn7. \textbf{O. GEA} means explanation accuracy on original graph and \textbf{P. GEA} indicates explanation accuracy on perturbed graph. Moreover, \textbf{$\Delta$GEA} is the explanation accuracy change after perturbation, \textbf{Sim}$_{\cos}$ is the cosine similarity between explanations (before and after perturbations), and \textbf{\#Pert.} represents the number of flipped edges. Besides, \textbf{$\Delta$Label} is the average change of predicted labels, while \textbf{$\Delta$Prob} denotes the average absolute change of predicted probability (of the original predicted class).  ``G5" is the set of nodes of which the original prediction confidence located in [0.5,0.6). Similarly, ``G6" for [0.6,0.7), ``G7" for [0.7,0.8), ``G8" for [0.8,0.9), ``G9" for [0.9,1.0], and ``All" for all nodes.}
	\centering
 \resizebox{\linewidth}{!}{
	\begin{tabular}{cccccccc}
		\toprule
		\textbf{Dataset} & \textbf{Confidence Group} & G5  & G6 & G7 & G8 & G9 & All\\
		\midrule

        \multirow{6}{*}{\rotatebox{90}{Syn1}} 
        & \textbf{O. GEA} & 0.696 & 0.763 & 0.726 & 0.775 & 0.683 & 0.728\\
        & \textbf{P. GEA} $(\downarrow)$ & 0.403 & 0.463 & 0.485 & 0.584 & 0.679 & 0.560\\
        & \textbf{$\Delta$GEA} $(\uparrow)$ & 0.293 & 0.299 & 0.241 & 0.190 & 0.004 & 0.168\\
        & \textbf{Sim}$_{\cos}$ $(\downarrow)$ & 0.817 & 0.868 & 0.898 & 0.937 & 0.987 & 0.923\\
        & \textbf{\#Pert.} $(\downarrow)$ & 6.5 & 5.0 & 4.5 & 2.7 & 0.2 & 3.0\\
        & \textbf{$\Delta$Label} $(\downarrow)$ & 0 & 0 & 0 & 0 & 0 & 0\\
        & \textbf{$\Delta$Prob} $(\downarrow)$ & 0.327 & 0.204 & 0.147 & 0.065 & 0.002 & 0.108\\
        \hline

        \multirow{6}{*}{\rotatebox{90}{Syn2}} 
        & \textbf{O. GEA} & 0.694 & 0.739 & 0.690 & 0.694 & 0.546 & 0.648\\
        & \textbf{P. GEA} $(\downarrow)$ & 0.329 & 0.396 & 0.390 & 0.387 & 0.381 & 0.347\\
        & \textbf{$\Delta$GEA} $(\uparrow)$ & 0.365 & 0.293 & 0.300 & 0.307 & 0.165 & 0.269\\
        & \textbf{Sim}$_{\cos}$ $(\downarrow)$ & 0.958 & 0.960 & 0.971 & 0.979 & 0.991 & 0.977\\
        & \textbf{\#Pert.} $(\downarrow)$ & 7.1 & 6.6 & 6.2 & 6.3 & 3.8 & 5.6\\
        & \textbf{$\Delta$Label} $(\downarrow)$ & 0 & 0 & 0 & 0 & 0 & 0\\
        & \textbf{$\Delta$Prob} $(\downarrow)$ & 0.353 & 0.268 & 0.181 & 0.103 & 0.003 & 0.143\\
        \hline

        \multirow{6}{*}{\rotatebox{90}{Syn3}} 
        & \textbf{O. GEA} & 0.484 & 0.496 & 0.479 & 0.510 & 0.440 & 0.480\\
        & \textbf{P. GEA} $(\downarrow)$ & 0.305 & 0.303 & 0.279 & 0.335 & 0.310 & 0.306\\
        & \textbf{$\Delta$GEA} $(\uparrow)$ & 0.179 & 0.193 & 0.200 & 0.175 & 0.130 & 0.179\\
        & \textbf{Sim}$_{\cos}$ $(\downarrow)$ & 0.966 & 0.973 & 0.962 & 0.981 & 0.989 & 0.978\\
        & \textbf{\#Pert.} $(\downarrow)$ & 8.0 & 7.4 & 6.6 & 5.8 & 4.7 & 6.7\\
        & \textbf{$\Delta$Label} $(\downarrow)$ & 0.008 & 0 & 0 & 0 & 0 & 0.002\\
        & \textbf{$\Delta$Prob} $(\downarrow)$ & 0.242 & 0.206 & 0.166 & 0.099 & 0.039 & 0.165\\
        \hline

        \multirow{6}{*}{\rotatebox{90}{Syn4}} 
        & \textbf{O. GEA} & 0.500 & 0.500 & 0.493 & 0.495 & 0.416 & 0.471\\
        & \textbf{P. GEA} $(\downarrow)$ & 0.309 & 0.314 & 0.319 & 0.326 & 0.296 & 0.311\\
        & \textbf{$\Delta$GEA} $(\uparrow)$ & 0.192 & 0.187 & 0.179 & 0.168 & 0.120 & 0.159\\
        & \textbf{Sim}$_{\cos}$ $(\downarrow)$ & 0.973 & 0.972 & 0.962 & 0.979 & 0.989 & 0.980\\
        & \textbf{\#Pert.} $(\downarrow)$ & 6.9 & 6.8 & 6.4 & 6.2 & 3.9 & 5.7\\
        & \textbf{$\Delta$Label} $(\downarrow)$ & 0 & 0 & 0 & 0 & 0 & 0\\
        & \textbf{$\Delta$Prob} $(\downarrow)$ & 0.333 & 0.267 & 0.177 & 0.111 & 0.030 & 0.149\\
        \hline

        \multirow{6}{*}{\rotatebox{90}{Syn5}} 
        & \textbf{O. GEA} & 0.456 & 0.470 & 0.443 & 0.461 & 0.430 & 0.450\\
        & \textbf{P. GEA} $(\downarrow)$ & 0.252 & 0.264 & 0.282 & 0.305 & 0.336 & 0.293\\
        & \textbf{$\Delta$GEA} $(\uparrow)$ & 0.204 & 0.207 & 0.161 & 0.157 & 0.094 & 0.157\\
        & \textbf{Sim}$_{\cos}$ $(\downarrow)$ & 0.967 & 0.971 & 0.975 & 0.979 & 0.984 & 0.976\\
        & \textbf{\#Pert.} $(\downarrow)$ & 7.5 & 7.6 & 7.0 & 6.6 & 5.2 & 6.6\\
        & \textbf{$\Delta$Label} $(\downarrow)$ & 0.013 & 0 & 0 & 0 & 0 & 0.002\\
        & \textbf{$\Delta$Prob} $(\downarrow)$ & 0.287 & 0.226 & 0.147 & 0.085 & 0.031 & 0.138\\
        \hline

        \multirow{6}{*}{\rotatebox{90}{Syn6}} 
        & \textbf{O. GEA} & 0.415 & 0.412 & 0.407 & 0.382 & 0.291 & 0.332\\
        & \textbf{P. GEA} $(\downarrow)$ & 0.260 & 0.249 & 0.239 & 0.259 & 0.217 & 0.231\\
        & \textbf{$\Delta$GEA} $(\uparrow)$ & 0.155 & 0.162 & 0.168 & 0.123 & 0.074 & 0.101\\
        & \textbf{Sim}$_{\cos}$ $(\downarrow)$ & 0.975 & 0.975 & 0.979 & 0.986 & 0.984 & 0.989\\
        & \textbf{\#Pert.} $(\downarrow)$ & 7.3 & 7.8 & 7.5 & 7.0 & 5.0 & 5.9\\
        & \textbf{$\Delta$Label} $(\downarrow)$ & 0 & 0 & 0 & 0 & 0 & 0\\
        & \textbf{$\Delta$Prob} $(\downarrow)$ & 0.348 & 0.270 & 0.189 & 0.106 & 0.029 & 0.092\\
        \hline

        \multirow{6}{*}{\rotatebox{90}{Syn7}} 
        & \textbf{O. GEA} & 0.428 & 0.415 & 0.406 & 0.405 & 0.401 & 0.409\\
        & \textbf{P. GEA} $(\downarrow)$ & 0.266 & 0.250 & 0.294 & 0.263 & 0.278 & 0.263\\
        & \textbf{$\Delta$GEA} $(\uparrow)$ & 0.161 & 0.165 & 0.156 & 0.142 & 0.123 & 0.146\\
        & \textbf{Sim}$_{\cos}$ $(\downarrow)$ & 0.986 & 0.983 & 0.989 & 0.991 & 0.992 & 0.990\\
        & \textbf{\#Pert.} $(\downarrow)$ & 8.2 & 7.9 & 7.5 & 7.1 & 6.1 & 7.2\\
        & \textbf{$\Delta$Label} $(\downarrow)$ & 0 & 0 & 0 & 0 & 0 & 0\\
        & \textbf{$\Delta$Prob} $(\downarrow)$ & 0.319 & 0.253 & 0.169 & 0.097 & 0.035 & 0.147\\
        \bottomrule
	\end{tabular}
 }
	\label{tab:PreConfidenceAll}
\end{table}

\end{document}